\pdfoutput=1

\documentclass[10pt,a4paper]{article}

\usepackage{lrec2022} 
\usepackage{multibib}
\newcites{languageresource}{Language Resources}
\usepackage{tabularx}
\usepackage{soul}

\usepackage{hyperref}

\usepackage{times}
\usepackage{latexsym}

\usepackage{danudefs}
\usepackage{algorithmic}
\usepackage{algorithm}
\usepackage{booktabs}
\usepackage{xspace}
\usepackage{url}
\usepackage{mathtools}
\usepackage{amsmath}
\usepackage{graphicx}
\usepackage{amsthm}
\usepackage{multirow}
\usepackage{enumerate}
\usepackage{esvect}
\usepackage{csvsimple}
\usepackage{microtype}
\usepackage{subcaption}
\usepackage{sidecap}

\usepackage{titlesec}
\titleformat{\section}{\normalfont\large\bfseries\center}{\thesection.}{1em}{}
\titleformat{\subsection}{\normalfont\SmallTitleFont\bfseries\raggedright}{\thesubsection.}{1em}{}
\titleformat{\subsubsection}{\normalfont\normalsize\bfseries\raggedright}{\thesubsubsection.}{1em}{}
\renewcommand\thesection{\arabic{section}}
\renewcommand\thesubsection{\thesection.\arabic{subsection}}
\renewcommand\thesubsubsection{\thesubsection.\arabic{subsubsection}}

\usepackage{epstopdf}
\usepackage[utf8]{inputenc}

\usepackage{array}
\newcolumntype{H}{>{\setbox0=\hbox\bgroup}c<{\egroup}@{}}

\usepackage[english]{babel}
\usepackage{hyperref}
\addto\captionsenglish{%
}
\addto\extrasenglish{%
}

\newtheorem{theorem}{Theorem}

\title{Query Obfuscation by Semantic Decomposition}

\name{$\textrm{Danushka Bollegala}^{1,2}$\thanks{Danushka Bollegala holds concurrent appointments as a Professor at University of Liverpool and as an Amazon Scholar. This paper describes work performed at the University of Liverpool and is not associated with Amazon.}, $\textrm{Tomoya Machide}^{3}$, $\textrm{Ken-ichi Kawarabayashi}^{3}$} 

\address{$\textrm{University of Liverpool}^{1}$, $\textrm{Amazon}^2$, $\textrm{National Institute of Informatics}^{3}$ \\
         danushka@liverpool.ac.uk, machide@nii.ac.jp, k\_keniti@nii.ac.jp}

\abstract{
We propose a method to protect the privacy of search engine users by decomposing the queries using
semantically \emph{related} and unrelated \emph{distractor} terms. Instead of a single query, the search engine
receives multiple decomposed query terms. Next, we reconstruct the search results relevant to the original
query term by aggregating the search results retrieved for the decomposed query terms.
We show that the word embeddings learnt using a distributed representation learning method can be used to find semantically related and distractor query terms.
We derive the relationship between the \emph{obfuscity} achieved through the proposed query anonymisation method and the \emph{reconstructability} of the original search results using the decomposed queries.
We analytically study the risk of discovering the search engine users' information intents under the proposed
query obfuscation method, and empirically evaluate its robustness against clustering-based attacks.
Our experimental results show that the proposed method can accurately reconstruct the search results for user queries, without compromising the privacy of the search engine users.
 \\ \newline  }

\begin{document}

\maketitleabstract

\section{Introduction}
\label{sec:introduction}

As web search engine users, we are left with two options regarding our privacy.
First, we can trust the search engine not to disclose the keywords that we use in a search session to third parties,
or even to use for any other purpose other than providing search results to the users who issued the queries.
However, the user agreements in most web search engines do not allow such user rights.
Although search engines pledge to protect the privacy of their users by encrypting queries and search results,\footnote{\url{https://goo.gl/JSBvpK}} the encryption is between the user and the search engine -- the original non-encrypted queries are still available to the search engine. 
The keywords issued by the users are a vital source of information for improving the relevancy of the search engine
and displaying relevant adverts to the users. 
For example, in learning to rank~\cite{He:2008}, keywords issued by a user and the documents clicked by that user are recorded by the search engine to learn the optimal dynamic ranking of the search results, user interests and extract attributes related to frequently searched entities~\cite{pasca:2014:EMNLP2014,Sadikov:WWW:2010,Santos:WWW:2010,Richardson:TWEB:2008,Pasca:WWW:2007}. 
Considering the fact that placing advertisements for the highly bid keywords is one of the main revenue sources for search engines,
there are obvious commercial motivations for the search engines to exploit the user queries beyond simply providing relevant search results to their users. 
For example, it has been reported that advertisements contribute to 96\% of Google's revenue.\footnote{\url{https://www.wordstream.com/articles/google-earnings}}
Therefore, it would be unrealistic to assume that the user queries will not be exploited in a manner unintended by the users.

\begin{figure}[t]
\centering
\includegraphics[width=70mm]{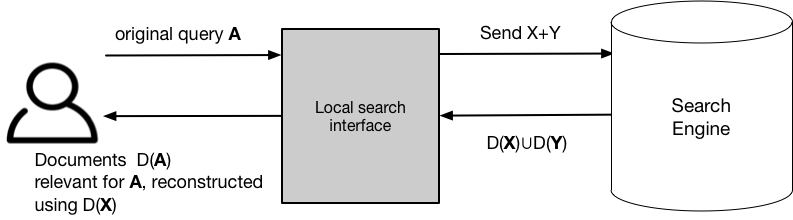}
\caption{Overview of the proposed method. The original query $A$ is decomposed into a set of relevant ($X$) and distractor ($Y$) terms
at the user-end. The search engine returns documents relevant for both $X$ and $Y$, denoted by $D(X) \cup D(Y)$. We will ignore $D(Y)$ and
reconstruct the search results for $A$ using $D(X)$.}
\label{fig:overview}
\end{figure}

As an alternative approach that does not rely on the goodwill of the search engine companies, we propose a method (shown in Figure~\ref{fig:overview}), where we disguise the queries that are sent to a search engine such that it is difficult for the search engine to guess the real information need of the user by looking at the keywords, yet it is somehow possible for the users to \emph{reconstruct} the search results relevant for them from what is returned by the search engine. 
The proposed method does not require any encryption or blindly trusting the search engine companies or any third-party mediators.
However, this is a non-trivial task because a search engine must be able to recognise the information need of a user in order to provide relevant results in the first place. 
Therefore, query obfuscation and relevance of search results are at a direct trade-off.

Specifically, given a user query $A$, our proposed method first finds a set of $n$ noisy relevant terms $X_1, X_2, \ldots, X_n$ (denoted by the set $\{X_i\}_{i=1}^{n}$)
and $m$ distractor terms $Y_1, Y_2, \ldots, Y_m$ (denoted by the set $\{Y_j\}_{j=1}^{n}$) for $A$ to obfuscate the user query. 
We use pre-trained word embeddings for identifying the noisy-relevant and distractor terms. 
We add Gaussian noise to the relevant terms such that it becomes difficult for the search engine to discover $A$ using $\{X_i\}_{i=1}^{n}$. 
However,  $\{X_i\}_{i=1}^{n}$ is  derived using $A$, so there is a risk that the search engine will perform some form of de-noising to unveil $A$ from $\{X_i\}_{i=1}^{n}$. 
Therefore, using $\{X_i\}_{i=1}^{n}$ alone as the keywords does not guarantee obfuscity.
To mitigate this risk, we generate a set of distractor terms   $\{Y_j\}_{j=1}^{n}$ separately for each user query.
We then issue $X_1, X_2, \ldots, X_n, Y_1, Y_2, \ldots, Y_m$ in random order to the search engine to retrieve the corresponding search results.
We then reconstruct the search results for $A$ using the search results we retrieve from the noisy-relevant terms and discard the search results retrieved from the distractor terms. 
It is noteworthy that during any stage of the proposed method,  we \emph{do not} issue $A$ as a standalone query nor in conjunction with any other terms to the search engine.
 Moreover, we do not require access to the search index, which is typically not shared by the search engine companies with the outside world.
  
We do not collect, process or release any personal data with ethical considerations in this work.
Our contributions in this paper can be summarised as follows:
\begin{itemize}
\item We propose a method to obfuscate user queries sent to a search engine by semantic decomposition to protect the privacy of the search engine users. 
Our proposed method uses pre-trained word embeddings.
\item We introduce the concepts of \emph{obfuscity} (i.e., how difficult it is to guess the original user query by looking at the auxiliary queries
sent to the search engine?), and \emph{reconstructability} (i.e. how easy it is to reconstruct the search results for the original query from the search results for the auxiliary queries?), and propose methods to estimate their values.
\item We theoretically derive the relationship between obfuscity and reconstructability using known properties of distributed word representations.
\item We evaluate the robustness of the proposed query obfuscation method against clustering-based attacks, where a search engine would cluster the keywords it receives within a single session to filter out distractors and predict the original query from the induced clusters. 
Our experimental results show that by selecting appropriate distractor terms, it is possible to guarantee query obfuscity, while reconstructing the relevant search results.
\end{itemize}

\section{Query Obfuscation}

\subsection{Finding Noisy-Related Terms}
\label{sec:noise}

Expanding a user query using related terms~\cite{Carpineto:2012} is a popular technique in information retrieval to address sparse results. 
Although query expansion is motivated as a technique for improving recall, we take a different perspective in this paper -- we consider query expansion as a method for obfuscating~\cite{Gervais_2014} the search intent of a user. 
Numerous methods have been proposed in prior work on query expansion to find good candidate terms for expanding a given user query such as using pre-compiled thesauri containing related terms and query logs~\cite{Carpineto:2012}.
We note that any method that can find related terms for a given user query $A$ can be used for our purpose given that the following requirements are satisfied:
\begin{enumerate}
\item The user query $A$ must never be sent to the search engine when retrieving related terms for $A$ because this would obviously compromise the obfuscation goal.

\item Repeated queries to the search engine must be minimised in order to reduce the burden on the search engine.
We assume that the query obfuscation process to take place outside of the search engine using a publicly available search API.
Although modern Web search engines would gracefully scale with the number of users/queries, obfuscation methods
that send excessively large numbers of queries are likely to be banned by the search engines because of the processing overhead.
Therefore, it is important that we limit the search queries that we issue to  the search engine when computing the related terms. 

 \item No information regarding the distribution of documents nor the search index must be required by the related term
 identification method. If we had access to the index of the search engine, then we could easily find the terms that are
 co-occurring with the user query, thereby identifying related terms. However, we assume that the query obfuscation
 process happens outside of the search engine. None of the major commercial web search engines such as Google, Bing or Baidu
 provide direct access to their search indices due to security concerns. Therefore, it is realistic to assume that we will not
 have access to the search index during anytime of the obfuscation process, including the step where we find related terms
 to a given user query.
 
 \item The related terms must not be too similar to the original user query $A$ because that would enable the search engine
 to guess $A$ via the related terms it receives. For this purpose, we would add noise to the user query $A$ and find
 \emph{noisy related neighbours} that are less similar to $A$.
 \end{enumerate}
 
 We propose a method that uses pre-trained word embeddings  to find related terms for a user query that satisfy all of the above-mentioned requirements.
 Context-independent word embedding methods such as word2vec~\cite{Milkov:2013} and GloVe~\cite{Pennington:EMNLP:2014} can represent the meanings of words using low dimensional dense vectors. 
 Using word embeddings is also computationally attractive because they are low dimensional (typically $100-600$ dimensions are sufficient), consuming less memory and faster when computing similarity scores.
 Although we focus on single word queries for the ease of discussion, we note that by using context-sensitive phrase embeddings such as Elmo~\cite{Elmo} and BERT~\cite{BERT} we can obtain vectors representing multi-word queries, which we defer to future work.
  
 We denote the pretrained word embedding of a term $A$ by $v(A)$. 
 To perturbate word embeddings, we add a vector, $\vec{\theta} \in \R^{d}$,  sampled independently for each $A$ from the $d$-dimensional Gaussian  with a zero mean and a unit variance, and measure the cosine similarity between $v(A) + \vec{\theta}$  and each of the words $X_{i} \in \cV$ in a predefined and fixed vocabulary $\cV$, using their word embeddings $v(X_{i})$.
 We then select the top most similar words $\{X_i\}_{i=1}^{n}$ as the noisy related terms of $A$.
   
Let us denote the set of documents retrieved using a query $A$ by $\cD(A)$. 
If we use a sufficiently large number of related terms $X_i$ to $A$, we will be able to retrieve $\cD(A)$ exactly using 
\begin{equation}
 \label{eq:sum}
 \cD'(A) =  \bigcup_{i=1}^{n} \cD(X_i) .
\end{equation}
However, in practice we are limited to using a truncated list of $n$ related terms because of computational efficiency and to limit the number of queries sent to the search engine. Therefore, in practice $\cD'(A)$ will not be exactly equal to  $\cD(A)$.
Nonetheless, we assume the equality to hold in \eqref{eq:sum}, and later in the theoretical proofs given in the supplementary material discuss the approximation error.
To model the effect of ranking, we consider only the top-$\zeta$ ranked documents as $\cD(X_{i})$ and set $\zeta = 100$ in our experiments.

\subsection{Obfuscation via Distractor Terms}
\label{sec:unrel}

Searching using noisy related terms $X_i$ alone of a user query $A$, does not guarantee the obfuscity.
The probability of predicting the original user query increases with the number of related terms used.
Therefore, we require further mechanisms to ensure that it will be difficult for the search engine to predict $A$ from the queries it has seen.
For this purpose, we select a set of unrelated terms  $\{Y_j\}_{j=1}^{n}$, which we refer to as the \emph{distractor} terms.

Several techniques can be used to find the distractor terms for a given query $A$.
For example, we can randomly select terms from the vocabulary $\cV$ as the distractor terms.
However, such randomly selected distractor terms are unlikely to be coherent, and could be easily singled-out  from the related terms by the search engine.
If we know the semantic category of $A$ (e.g. $A$ is a \emph{person} or a \emph{location} etc.), then we can limit the distractor terms to the same semantic category as $A$. 
This will guarantee that both related terms as well as distractor terms are semantically related in the sense that they both represent the same category. 
Therefore, it will be difficult for the search engine to discriminate between related terms and distractor terms. 
Information about the semantic categories of terms can be obtained through different ways such as Wikipedia category pages, taxonomies such as the WordNet~\cite{WordNet} or by named entity recognition (NER) tools.
Moreover, we consider distractor terms $Y_{j}$ that have similar average frequency as the original query $A$ and the noisy related terms $X_{i}$ so that it will be difficult to differentiate between distractor terms and noisy related terms based on frequency information.

\begin{figure}
\centering
\includegraphics[width=50mm]{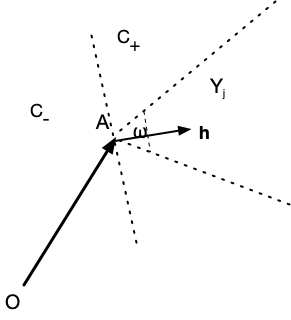}
\caption{Selecting distractor terms for a given query $A$. We first compute the noise ($\vec{\theta}$) added vector $\vec{A'}$ for $\vec{A}$, and then search for terms $Y_{j}$ that are located inside a cone that forms an angle $\omega$ with $\vec{A}'$. This would ensure that distractor terms are sufficiently similar to the noise component, therefore  difficult to distinguish from $A$.}
\label{fig:distselect}
\end{figure}

We propose a method to find distractor terms $Y_{j}$ for each query $A$  using pre-trained word embeddings as  illustrated in Figure~\ref{fig:distselect}.
Let us consider a set of candidate terms $\cC$ from which we must select the distractor terms.
For example, $\cC$ could be a randomly selected subset from the vocabulary of the corpus used to train word embeddings.
First, we select a random hyperplane (represented by the normal vector $\vec{h} \in \R^{d}$ to the hyperplane) in the embedding space that passes through the point corresponding to $A$. 
Next, we split $\cC$ into two mutually exclusive sets $\cC_{+} = \{x : x \in \cC, \vec{x}\T\vec{h} \geq 0\}$ and
$\cC_{-} =  \{x : x \in \cC, \vec{x}\T\vec{h} < 0\}$ depending on which side of the hyperplane the word is located.
Let us define $\cC_{\max}$ and $\cC_{\min}$ to be respectively the larger and smaller of the two sets $\cC_{+}$ and $\cC_{-}$
(i.e. $\cC_{\max} = \argmax_{\cS \in \{\cC_{+}, \cC_{-}\}} |\cC|$ and $\cC_{\min} = \argmin_{\cS \in \{\cC_{+}, \cC_{-}\}} |\cC|$)
 Next, we remove the top $10\%$ of the similar words in $\cC_{\max}$ to the original query $A$.
We then use this reduced $\cC_{\max}$ as $\cC$ (i.e. $\cC \leftarrow \cC_{\max}$) and 
repeat this process until we are left with the desired number of distractor terms in $\cC$.
Intuitively, we are partitioning the candidate set into two groups in each iteration considering some attribute (dimension) of the word embedding of the query (possibly representing some latent meaning of the query), and removing similar terms in that subspace.

\subsection{Reconstructing Search Results}
\label{sec:construct}

Once we have identified a set of noisy related terms, $\{X_i\}_{i=1}^{n}$,
and a set of distractor terms,  $\{Y_j\}_{j=1}^{n}$, we issue those terms as queries to the search engine
and retrieve the relevant search results for each individual term. 
We issue related terms and distractor terms in a random sequence, and ignore the results returned by the search engine for the distractor terms. 
Finally, we can reconstruct the search results for $A$ using \eqref{eq:sum}.

\section{Obfuscity vs. Reconstructability}
\label{sec:connection}

Our proposed query decomposition method strikes a fine balance between
two factors (a) the difficulty for the search engine to guess the original user query $A$, from the set of terms that it receives
$\cQ(A) = \{X_{1}, X_{2}, \ldots, X_{n}, Y_{1}, Y_{2}, \ldots, Y_{m}\}$, and 
(b) the difficulty to reconstruct the search results, $\cD(A)$, for the original user query, $A$, using the search results
for the noisy related terms following \eqref{eq:sum}.
We refer to (a) as the \emph{obfuscity}, and (b) as the \emph{reconstructability}
of the proposed query decomposition process. 

\subsection{Obfuscity}
\label{sec:obfuscity}

We define \emph{obfuscity}, $\alpha$, as the ease to guess the user query $A$, from the
terms issued to the search engine and compute it as follows:
\begin{equation}
 \label{eq:anon}
\alpha = 1- \frac{1}{|\cQ(A)|} \sum_{q \in \cQ(A)} \textrm{sim}(v(A), v(q))
\end{equation}
Specifically, we measure the average cosine similarity between the word embedding, $v(A)$, for the original user query $A$,
and the word embeddings $v(q)$ for each of $q \in \cQ(A)$ search terms.
If the similarity is higher, then it becomes easier for the search engine to guess $A$ from the search terms.
The difference between this average similarity and $1$ (i.e. the maximum value for the average similarity) is considered as a measure of obfuscity we can guarantee through the proposed query decomposition process.
Even if we are not exactly sending $A$ to the search engine as a keyword, the search engine will be able to figure out $A$ from $\cQ(A)$.
By using word embeddings to measure the similarity between the original query $A$ and the keywords $\cQ(A)$ sent to the search engine, we are able to consider not only exact matches but semantically similar keywords, which can be seen as a soft match between words.
The definition of obfuscity given by \eqref{eq:anon} is based on this intuition. 

\subsection{Reconstructability}
\label{sec:recon}

We reconstruct the search results for $A$ using the search results for the queries $\{X_i\}_{i=1}^{n}$ following \eqref{eq:sum}.
We define \emph{reconstructability}, $\rho$ as a measure of the accuracy of this reconstruction process and is defined as follows:
\begin{equation}
 \label{eq:recon}
\rho = \frac{|\cD(A) \cap \cD'(A)|}{|\cD(A)|}
\end{equation}
A document retrieved and ranked at top-$\zeta$ by only a single noisy related term might not be relevant to the original user query $A$.
A more robust reconstruction procedure would be to consider a document as relevant if it has been retrieved by 
at least $l$ different noisy related terms. If a user query $A$ can be represented by a set of documents where,
each document is retrieved by at least $l<n$ different noisy related terms, then we say $A$ to be \emph{$l$-reconstructable}.
In fact, the reconstruction process defined in \eqref{eq:sum} corresponds to the special case where $l = 1$.
Increasing the value of $l$ would decrease the number of relevant documents retrieved for the original user query $A$,
but it is likely to increase the relevance of the retrieval process.
In the supplementary material, we prove that the trade-off relationship \eqref{eq:rel-k} holds between $\rho$ and $\alpha$.
\begin{theorem}
Given a query $A$, represented by $d$-dimensional embedding, $v(A)$, let us obfuscate it with $n$ distractor terms and
use all (i.e. $n=l$) distractor terms to reconstruct the search results for $A$. 
The obfuscity $\alpha$ and the reconstructability $\rho$ is in the inverse  (trade-off) relationship given by \eqref{eq:rel-k}, where $c$ and $Z$
are query independent constants. 
\begin{align}
\label{eq:rel-k}
\log \rho = \frac{cl}{2d} \left( c + 2(1-\alpha)\norm{v(A)}_{2} \right) - \log Z &
\end{align}
\end{theorem}

\subsection{Extension to Multi-word Expressions}
The anonymisation method and its theoretical analysis described in the paper so far can be easily generalised to handle multi-word queries.
Specifically, in the case of multi-word queries we must embed not only unigrams but phrasal $n$-grams.
Directly modelling $n$-gram co-occurrences is challenging for higher-order $n$-grams because of data sparseness issues~~\cite{Turney:JAIR:2010}.
Compositional approaches~\cite{cordeiro-EtAl:2016:P16-1,Hashimoto:ACL:2016,Poliak:EACL:2017,TACL586} have been proposed to overcome this problem, where unigram, subword, or character level embeddings are iteratively combined to create representations for longer phrasal queries. 
These methods can compute length-invariant vector representations for $n$-grams, which can then be used in the same manner as described in Section~\ref{sec:noise} for finding noisy-related terms and in Section~\ref{sec:construct}  for finding distractor terms.

\subsection{Effect of Ranking}
If the number of documents containing $q$, $|\cD(q)|$, is less than $\zeta$ for all $q \in \cQ(A)$, we will be able to retrieve all documents containing the related and distractor terms.
However, when this condition does not hold for one or more terms in $\cQ(A)$, the reconstruction process is not guaranteed to perfectly reconstruct $\cD(A)$, depending on the accuracy of the ranking method used in the search engine.
Note that due to the relatedness between the terms $\{X_{i}\}_{i=1}^{n}$, even though a particular relevant document $d \in \cD(A)$ is not retrieved by a term $X_{i}$ due to the truncation by ranking, it could still be retrieved by a different $X_{j}$ ($j \neq i$) term.
Moreover, in practice, the number of relevant documents for a query is significantly smaller than $\zeta$ and modern search engines have accurate ranking models that return relevant results among top-$\zeta$, thus mitigating this risk of truncation.

In addition to reconstructing the search results for the original query from the search results for its related terms, we must also determine the ranked order of those search results in real-world settings, where potentially a large number of relevant results do exist for a given user query and simply determining only the match set is inadequate.
In the case of static ranking scores such as PageRank, which are independent of the query, we can use them to induce a total ordering in the reconstructed search results set.
Although dynamic rank scores might be available for the set of search results retrieved for each related term, it is not obvious how to compare rank scores for search results obtained for \emph{different} queries.
We do not consider the problem of ranking the reconstructed search results in this paper.

\section{Experiments}

\subsection{Effect of Noise and Distractor Terms}

\begin{figure*}[t]
    \centering
    \begin{minipage}{0.33\textwidth}
    	\centering
     		\includegraphics[width=44mm]{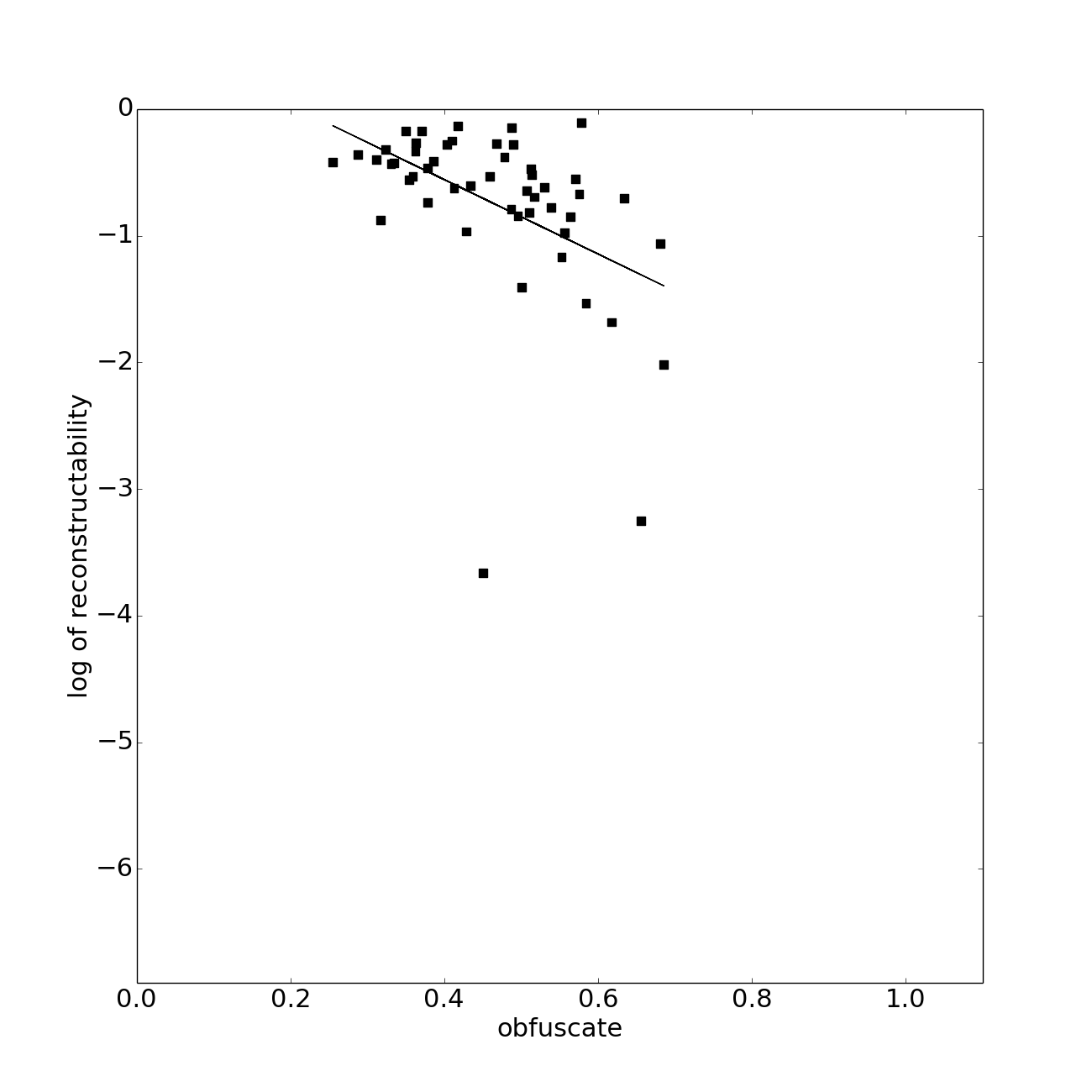}
	\end{minipage}%
 	\begin{minipage}{0.33\textwidth}
   		\centering
   		\includegraphics[width=44mm]{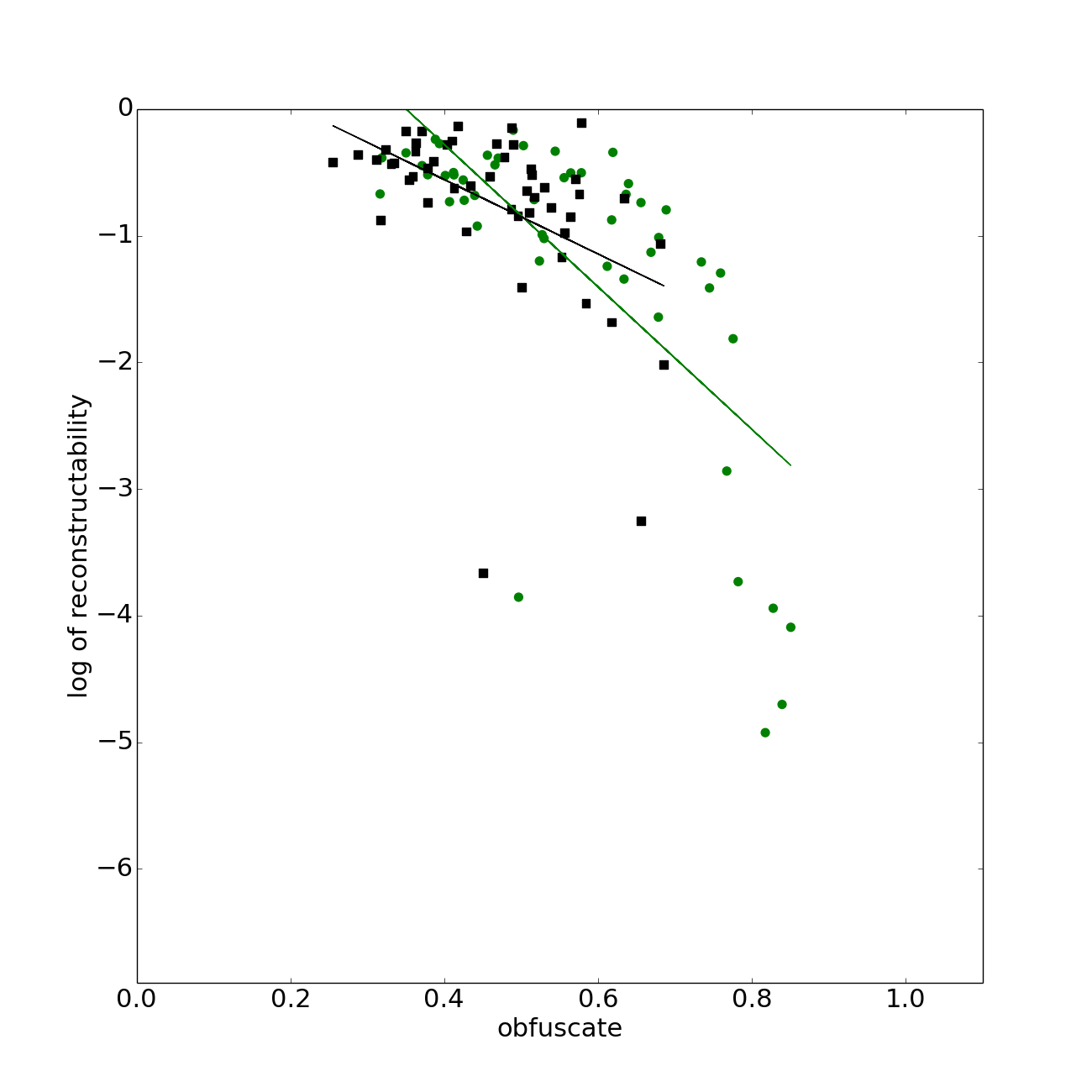}
	 \end{minipage}%
	\begin{minipage}{0.33\textwidth}
    		\centering
     		\includegraphics[width=44mm]{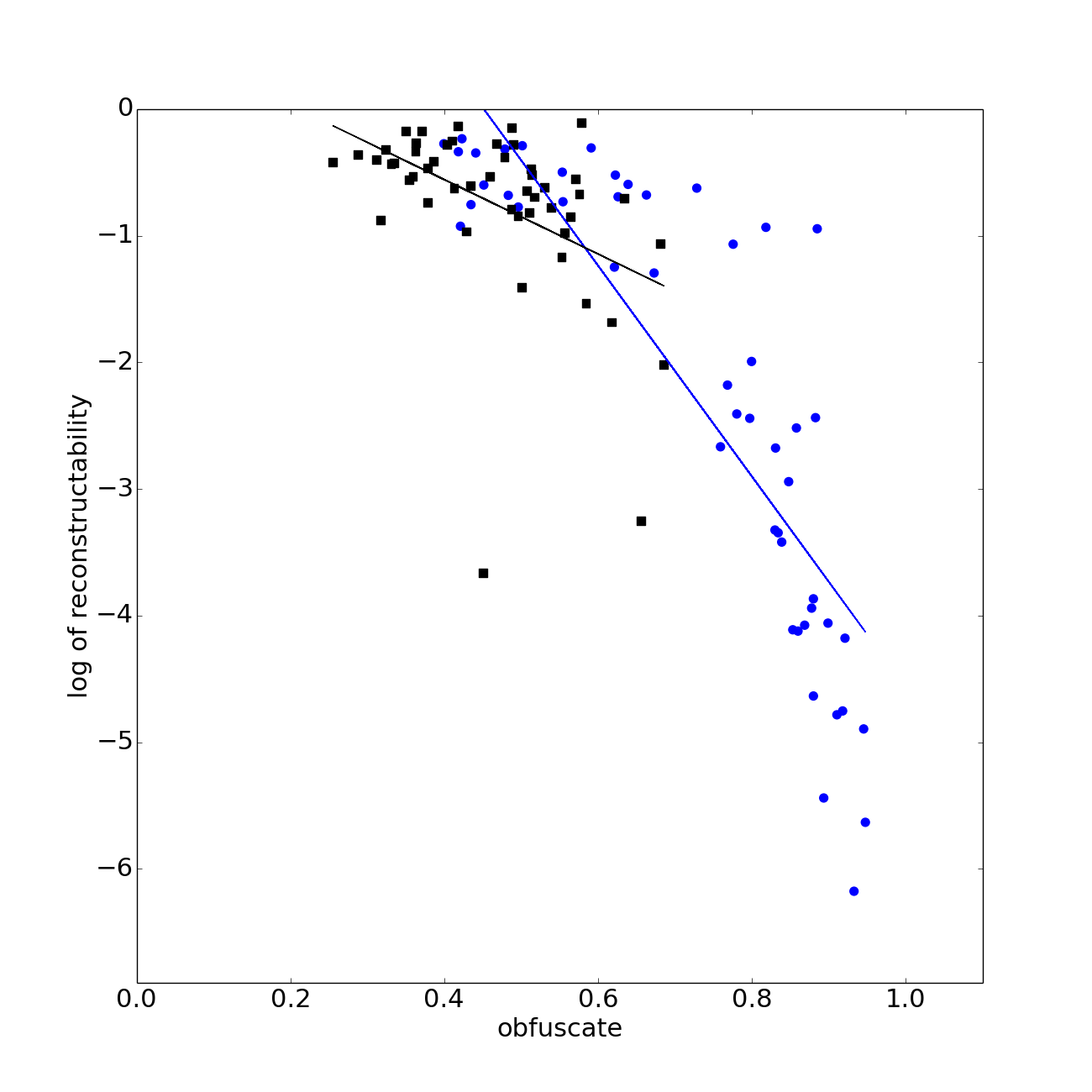}
 	\end{minipage}
	\caption{Relationship between obfuscity and reconstructability under different levels of added noise and no distractor terms (left: no-noise, middle: medium-level of noise, and right: high-level of noise). Base reconstructability scores for no noise and no distractor terms are super imposed in black boxes.}
	\label{fig:nodist}
\end{figure*} 

\begin{figure*}[t]
    \centering
    \begin{minipage}{0.33\textwidth}
    	\centering
     		\includegraphics[width=44mm]{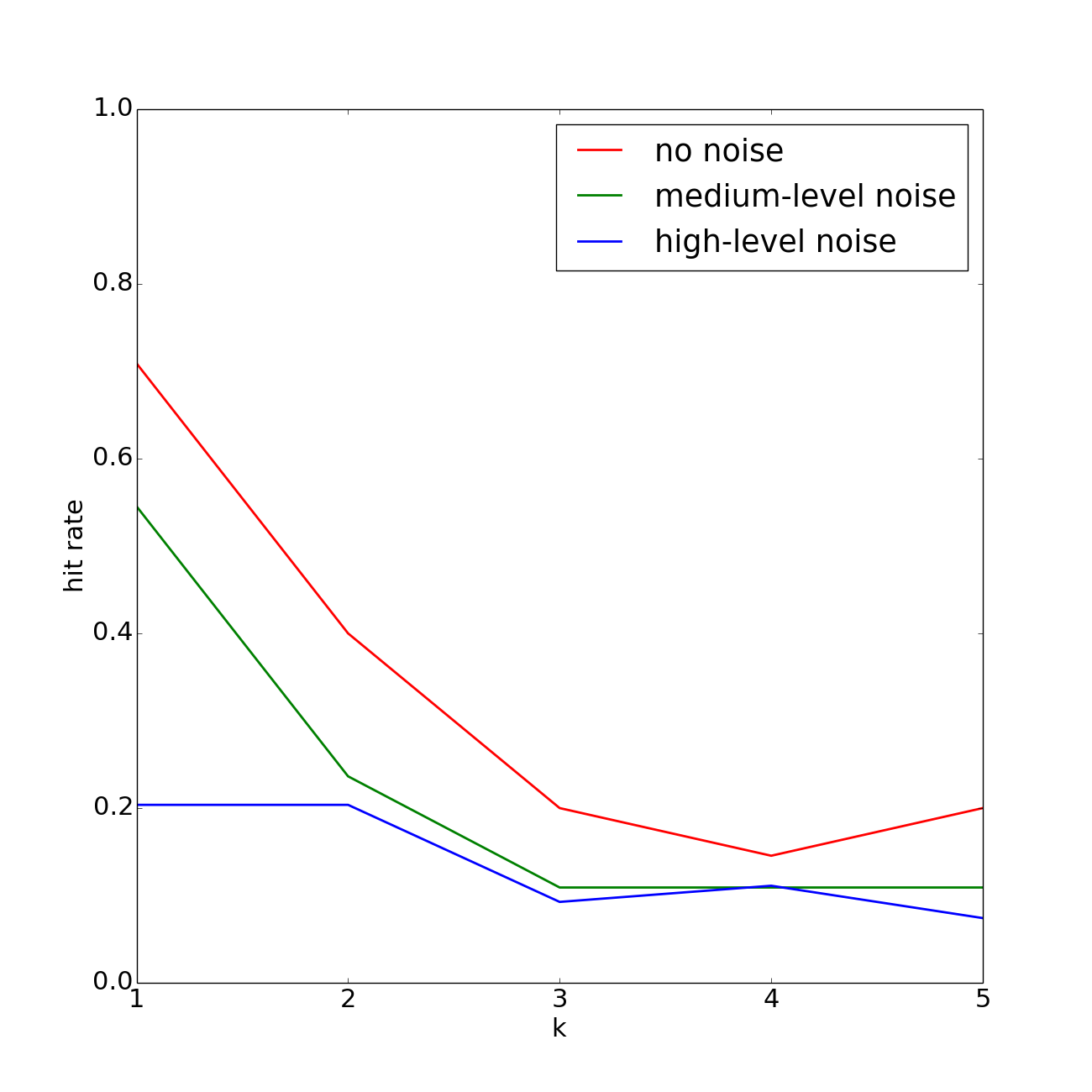}
	\end{minipage}%
 	\begin{minipage}{0.33\textwidth}
   		\centering
   		\includegraphics[width=44mm]{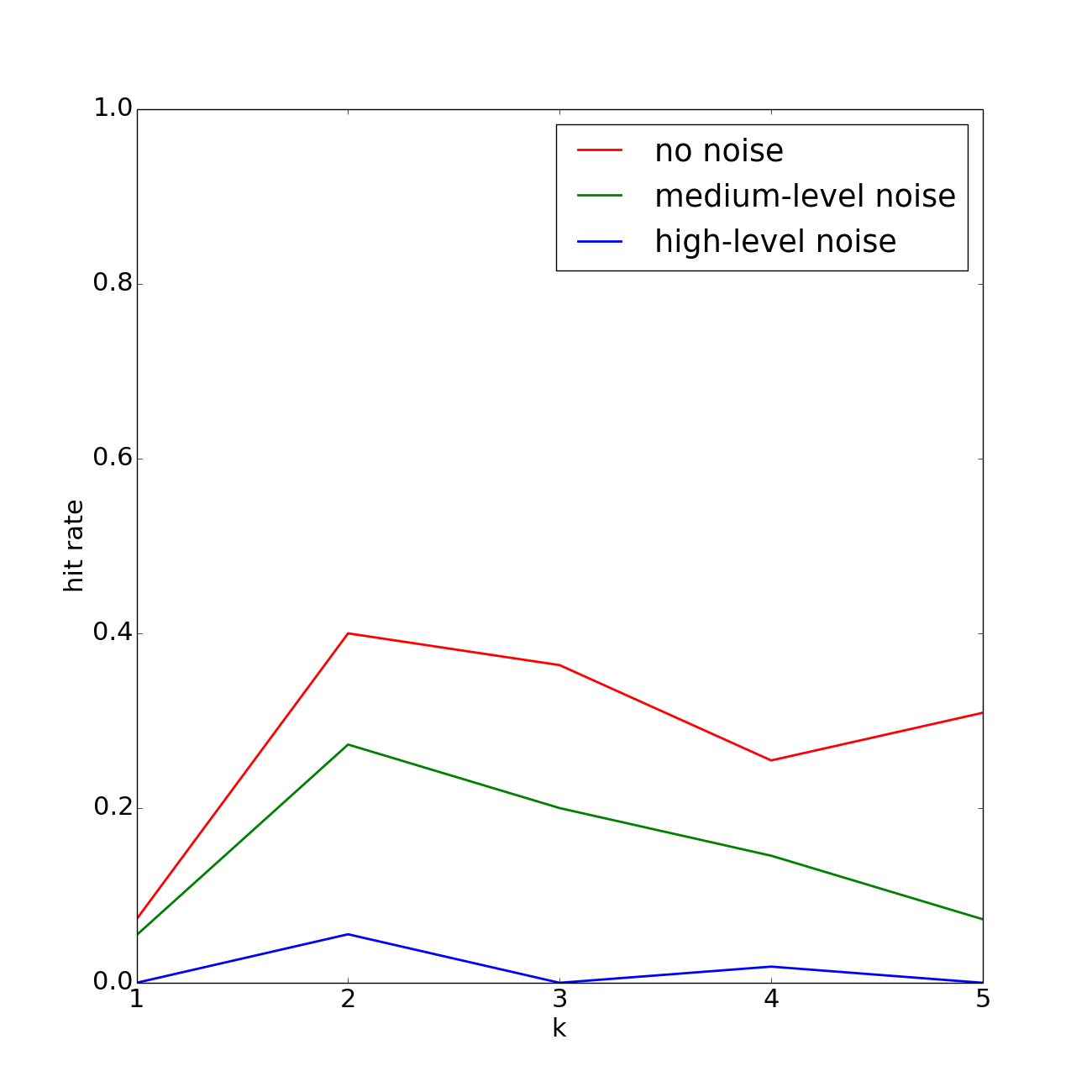}
	 \end{minipage}%
	\begin{minipage}{0.33\textwidth}
    		\centering
     		\includegraphics[width=44mm]{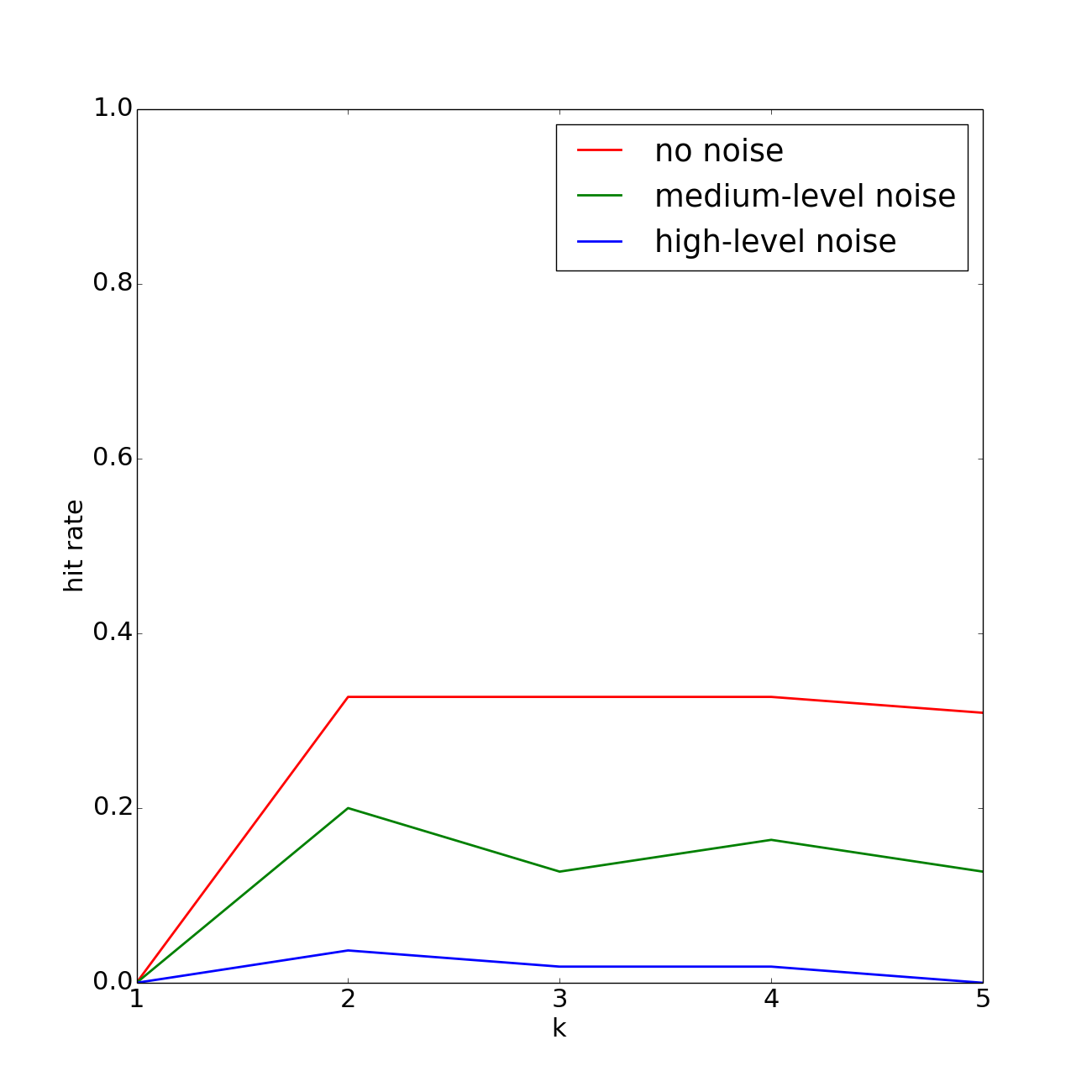}
 	\end{minipage}
	\caption{Hit rates for the $k$-means clustering attacks for increasing number of clusters ($k$) and distractor terms. (left: no distractors, middle: 20 distractors, and right: 40 distractors). In each figure, we show results for three levels of added noise.}
	\label{fig:clust}
\end{figure*}

\begin{figure*}[t]
    \centering
    \begin{minipage}{0.33\textwidth}
    	\centering
     		\includegraphics[width=44mm]{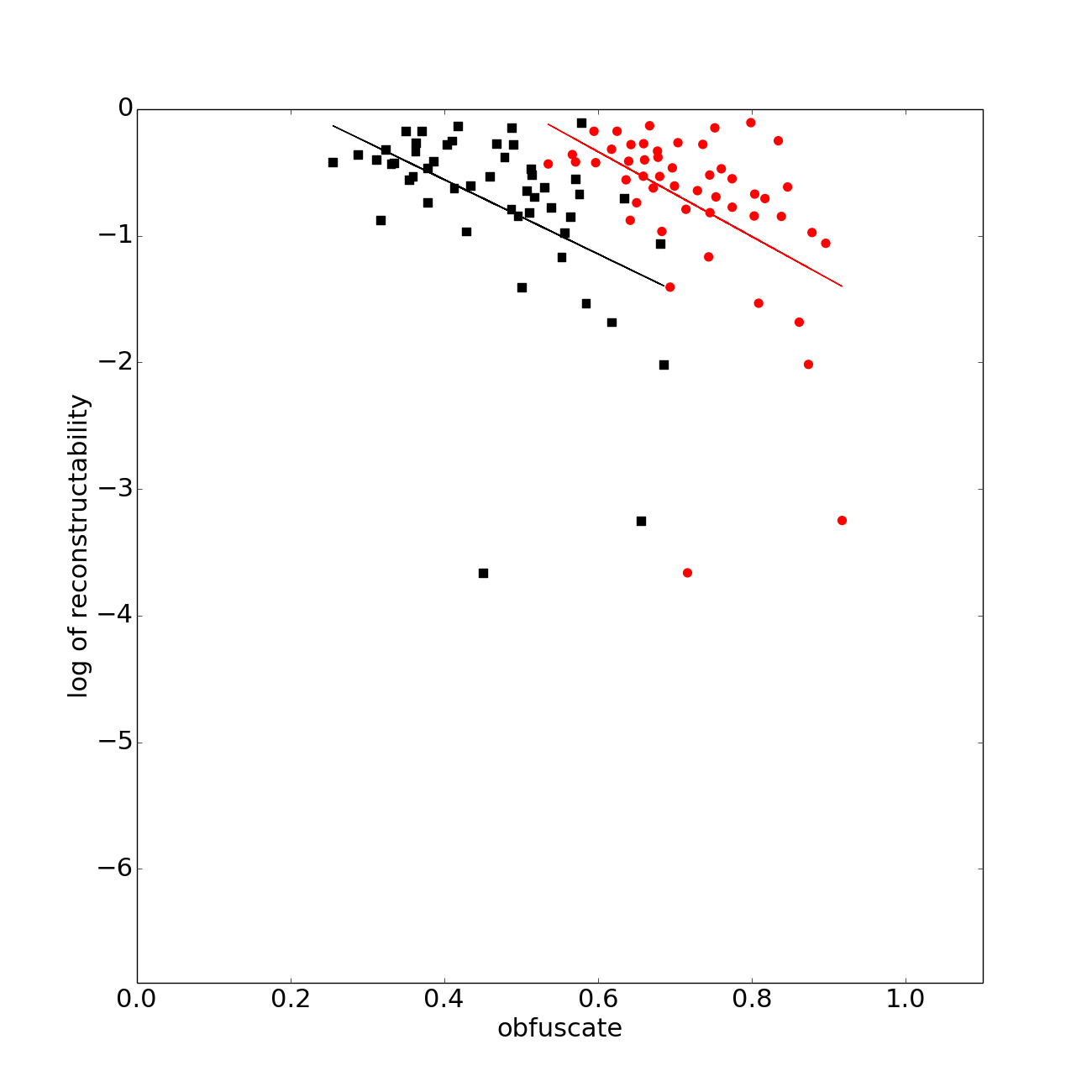}
	\end{minipage}%
 	\begin{minipage}{0.33\textwidth}
   		\centering
   		\includegraphics[width=44mm]{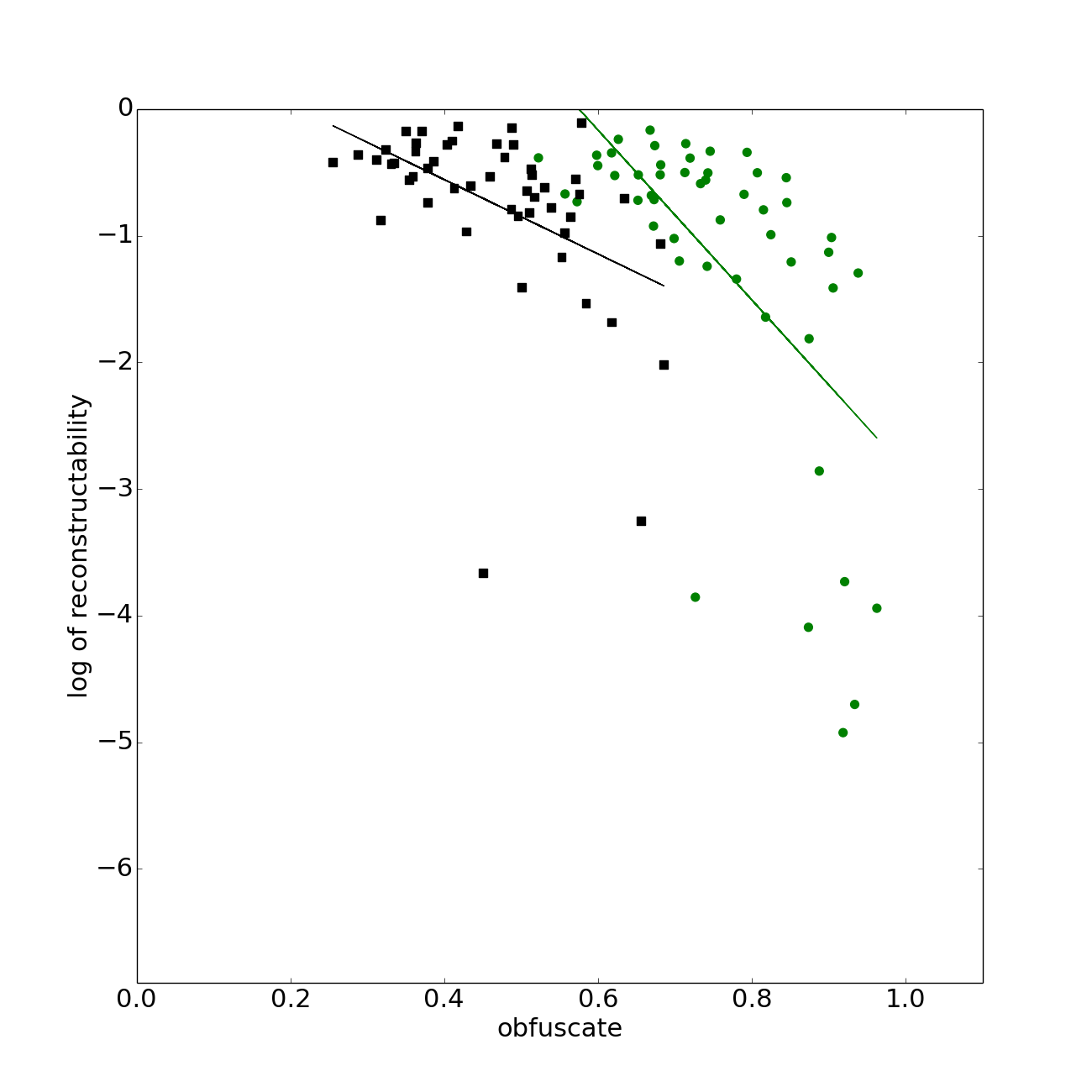}
	 \end{minipage}%
	\begin{minipage}{0.33\textwidth}
    		\centering
     		\includegraphics[width=44mm]{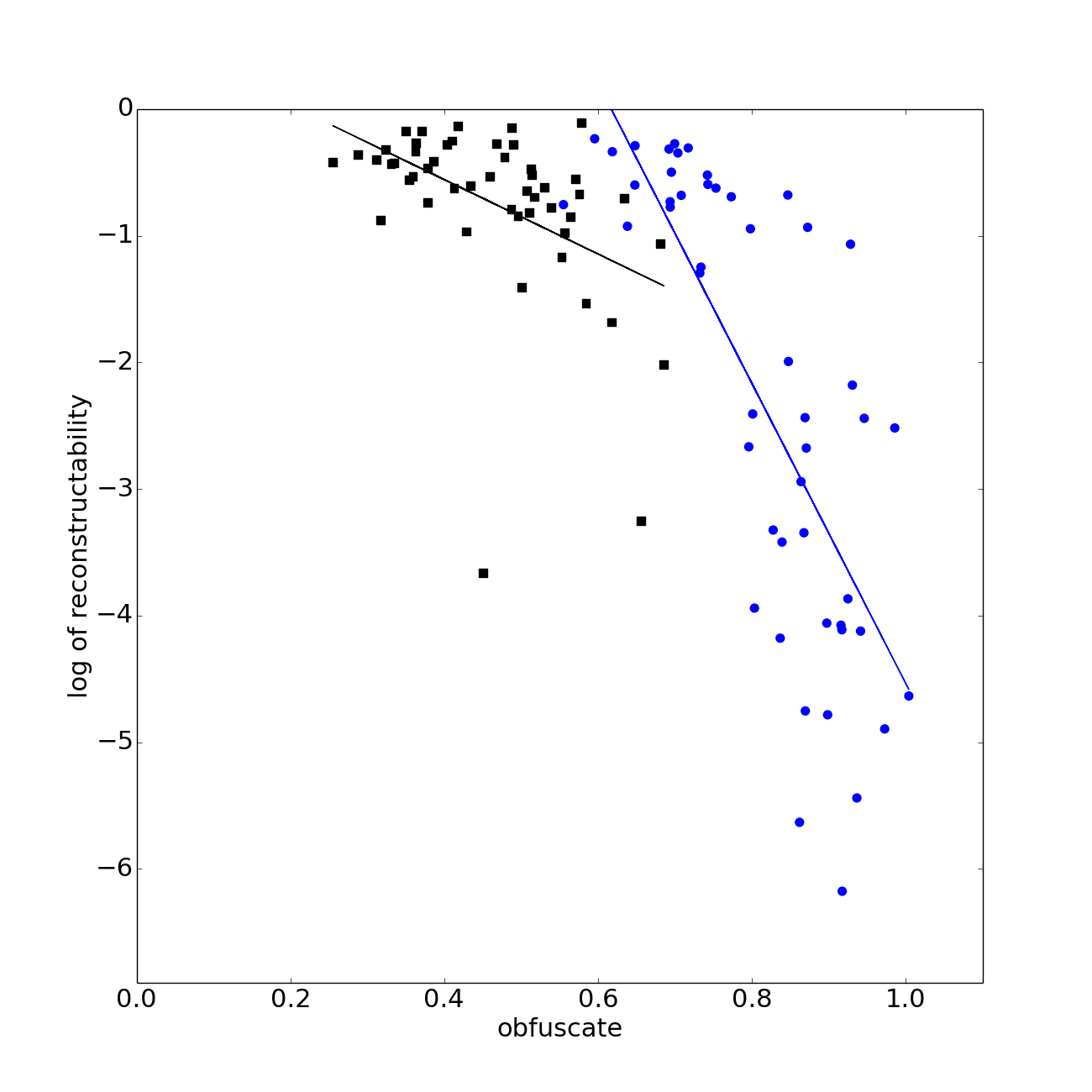}
 	\end{minipage}
	\caption{Relationship between obfuscity and reconstructability under different levels of added noise and with 20 distractor terms (left: no-noise, middle: medium-level of noise, and right: high-level of noise). Base reconstructability scores for no noise and no distractor terms are super imposed in black boxes.}
	\label{fig:20dist}
\end{figure*}

To evaluate the proposed method we create a dataset where we select $50$ popular queries from Wikipedia query logs and associate them with the relevant Wikipedia articles. 
We use the December 2015 dump of English Wikipedia for this purpose and build a keyword-based inverted search index.
We use 300 dimensional pretrained GloVe~\cite{Pennington:EMNLP:2014} embeddings trained from a 42 billion token Web crawled corpus.\footnote{\url{https://nlp.stanford.edu/projects/glove/}}
Figure~\ref{fig:nodist} and  show the obfuscity and the natural (base $e$) log-reconstructability to  values for the 50 queries in our dataset at different levels of noise. Specifically, we add Gaussian noise with zero-mean and standard deviations of 0.6 and 1.0 respectively to stimulate medium and high levels of noise, whereas the no-noise case corresponds to not perturbing the word embeddings. 
Trend with distractor terms are shown in \autoref{fig:20dist}.
Distractor terms that have similar average frequency to the original query and the noisy relevant terms are randomly selected from Wikipedia articles that belong to the same Wikipedia category tag as the article for the original query.

We see a negative correlation between obfuscity and reconstructability in all plots as predicted by \eqref{eq:rel-k}.
Addition of noise affects the selection of related terms but not the selection of distractor terms.
However, related terms influence both obfuscity as well as reconstructability. 
Because Gaussian noise is added to the word embedding of the original query, and the nearest neighbours to this noise added embedding are selected as the related terms, this process would help us to increase obfuscity.
On the other hand, the search results obtained using noisy related terms will be less relevant to the original user query.
Therefore, reconstructing the search results for the original user query using the search results for the noisy related terms will become more difficult, resulting in decreasing the reconstructability. 
The overall effect of increasing obfuscity and decreasing reconstructability is shown by the increased negative gradient of the line of best fit in the figures.

\subsection{Robustness against Attacks}
\label{sec:attack}

An important aspect of a query obfuscation method is its robustness against attacks.
Given that the proposed method sends two groups of terms (relevant and distractor) to a search engine, a natural line of attack is to cluster the received terms to filter out distractor terms and then guess the user query from the relevant terms.
We call such attacks as \emph{clustering attacks}.
As a concrete example, we simulate a clustering attacker who applies $k$-means clustering to the received terms. The similarity between terms for the purpose of clustering is computed using the cosine similarity between the corresponding word embeddings. Any clustering algorithm can be used for this purpose. We use $k$-means clustering because of its simplicity. Next, the attacker must identify a single cluster that is likely to contain the relevant terms. For this purpose, we measure the \emph{coherence}, $\mu(\cC)$, of a cluster $\cC$ given by \eqref{eq:coherence}.
\begin{align}
\label{eq:coherence}
\mu(\cC) = \frac{2}{|\cC|(|\cC|-1)}\sum_{u,v \in \cC,  u \neq v} \textrm{sim}(u,v)
\end{align}
Here, $u,v \in \cC$ are two distinct terms in $\cC$. Because a cluster containing relevant terms will be more coherent than a cluster containing distractor terms, the attacker selects the cluster with the highest coherence as the relevant cluster. Finally, we find the term from the entire vocabulary that is closest to the centroid of the cluster as the guess $\hat{A}$ of the original user query $A$.
We define \emph{hit rate} to be the proportion of the queries that we disclose via the clustering attack.
Figure~\ref{fig:clust} shows the hit rates for the clustering attacks under different numbers of distractor terms.

From Figure~\ref{fig:clust} left we see that the hit rate is high when we do not use any distractor terms. In this case, the set of candidate terms consists purely of related terms $X_{i}$. We see that if we cluster all the related terms into one cluster ($k = 1$) we can easily pick the original query $A$ by measuring the similarity to the centroid of the cluster. The hit rate drops when we add noise to the word embeddings, but even with the highest level of noise, we see that it is possible to discover the original query in 19\% of the time.
However, the hit rate drops significantly for all levels of noise when we add distractor terms as shown in the middle and right plots in Figure~\ref{fig:clust}. 

Hit rate is maximum when we set $k=2$, which is an ideal choice for the number of clusters considering the fact that we have two groups of terms (related terms and distractors) among the candidates. Increasing $k$ also increases the possibility of further splitting the related terms into multiple clusters thereby decreasing the probability of discovering the original query from a single cluster.
We see that hit rates under no or medium levels of noise drops when we increase the number of distractor terms from 20 to 40, but the effect on high-level noise added candidates is less prominent. 
This result suggests that we could increase the number of distractor terms while keeping the level of noise to a minimum. 

We show the terms discovered by clustering attacks for two example queries, \emph{Hitler}  (Table~\ref{tbl:example}) and \emph{mass murder} (Table~\ref{tbl:murder:10}) using a relatively small ($<10$) distractor terms.
 We see terms that are related to the original queries can be accurately identified from the word embeddings. 
Moreover, by adding a high-level of noise to the embeddings, we can generate distractor terms that are sufficiently further from the original queries.
 Consequently, we see that both obfuscity and reconstructability are relatively high for these examples. 
 Interestingly, the clustering attack is unable to discover the original queries, irrespective of the number of clusters produced.

\begin{table}[t]
\small
\centering
\begin{tabular}{p{25mm} p{40mm}}\toprule
Query & Hitler \\ \midrule
noise & high-level \\
related terms & nazi, f{\"u}hrer, gun, wehrmacht, guns, nra, pistol, bullets \\
obfuscity & 0.867 \\
reconstructability 	& 0.831 \\ \midrule
Clustering Attack & Revealed Query \\
k=1	& motagomery \\
k=2 & albany, george \\
k=3 & 	smith, albany \\
k=4	& smith, fresno \\
k=5	& rifle, albany \\
\bottomrule
\end{tabular}
\caption{Terms revealed by the clustering attacks for the query \emph{Hitler}. 
Clustering attack with different number of clusters ($k$) \emph{does not} reveal the original query.}
\label{tbl:example}
\vspace{-5mm}
\end{table}
\begin{table}[t]
\small
\centering
\begin{tabular}{p{25mm} p{40mm}}\toprule
Query & mass murder \\ \midrule
noise & high-level \\
related terms & terrorism, killed, wrath, full-grown\\
obfuscity & 0.789 \\
reconstructability 	&  0.747\\ \midrule
Clustering Attack & Revealed Query \\
k=1	& richmond\\
k=2 & fremont, death\\
k=4	&  pasadena, words\\
k=4	&  pasadena, words\\
k=5	&  pasadena, anderson\\
\bottomrule
\end{tabular}
\caption{Terms revealed by the clustering attacks for the query \emph{mass murder}. We see that the query nor its two tokens
are revealed by the clustering attacks with different $k$ values.}
\vspace{-5mm}
\label{tbl:murder:10}
\end{table}

\section{Trade-off between Reconstructability and the Hit Rate in Clustering Attacks}
\label{sec:tradeoff}

\begin{figure*}[t!]
	\centering
	\begin{subfigure}[t]{0.32\textwidth}
		\centering
		\includegraphics[width=44mm]{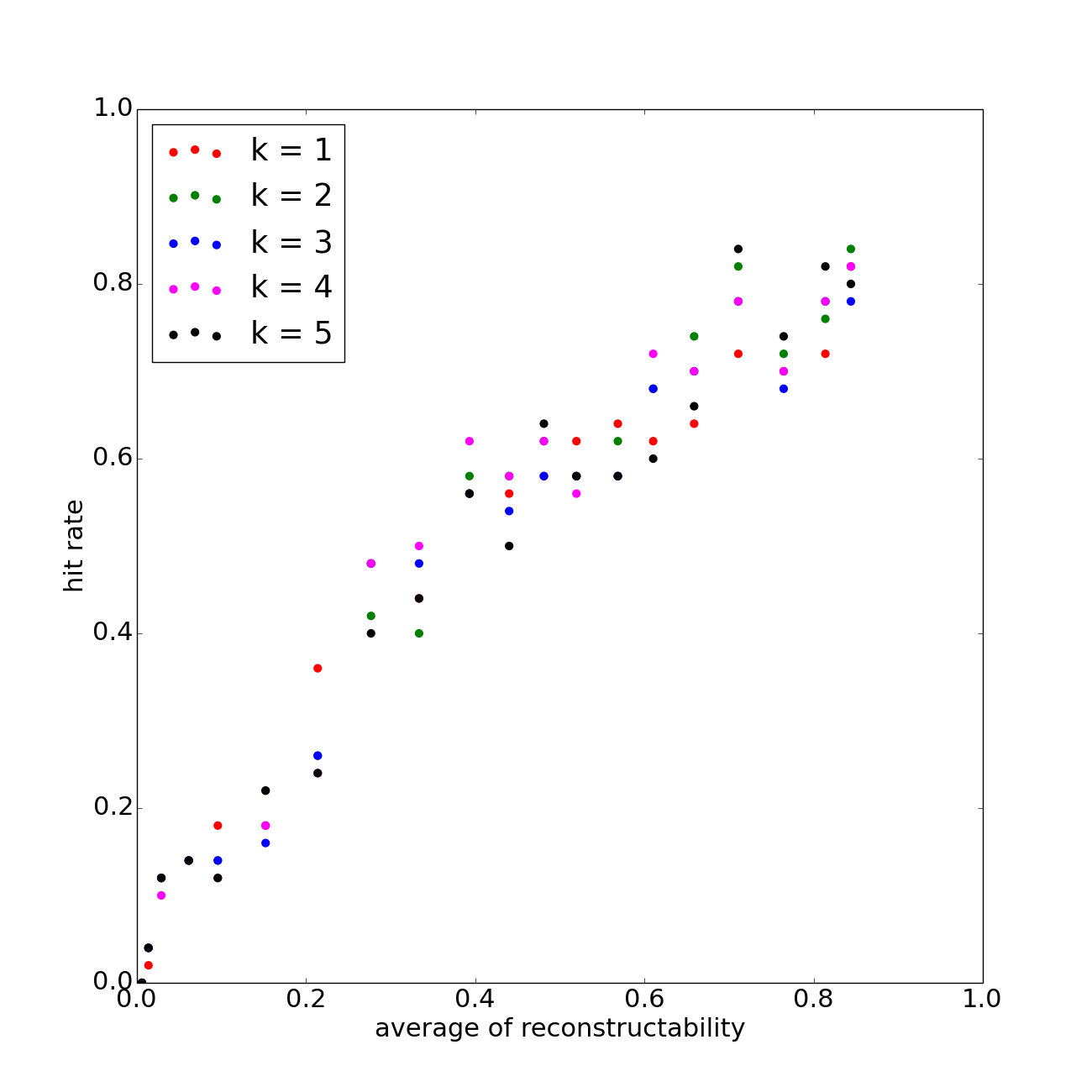}
	\end{subfigure}%
	~
	\begin{subfigure}[t]{0.32\textwidth}
		\centering
		\includegraphics[width=44mm]{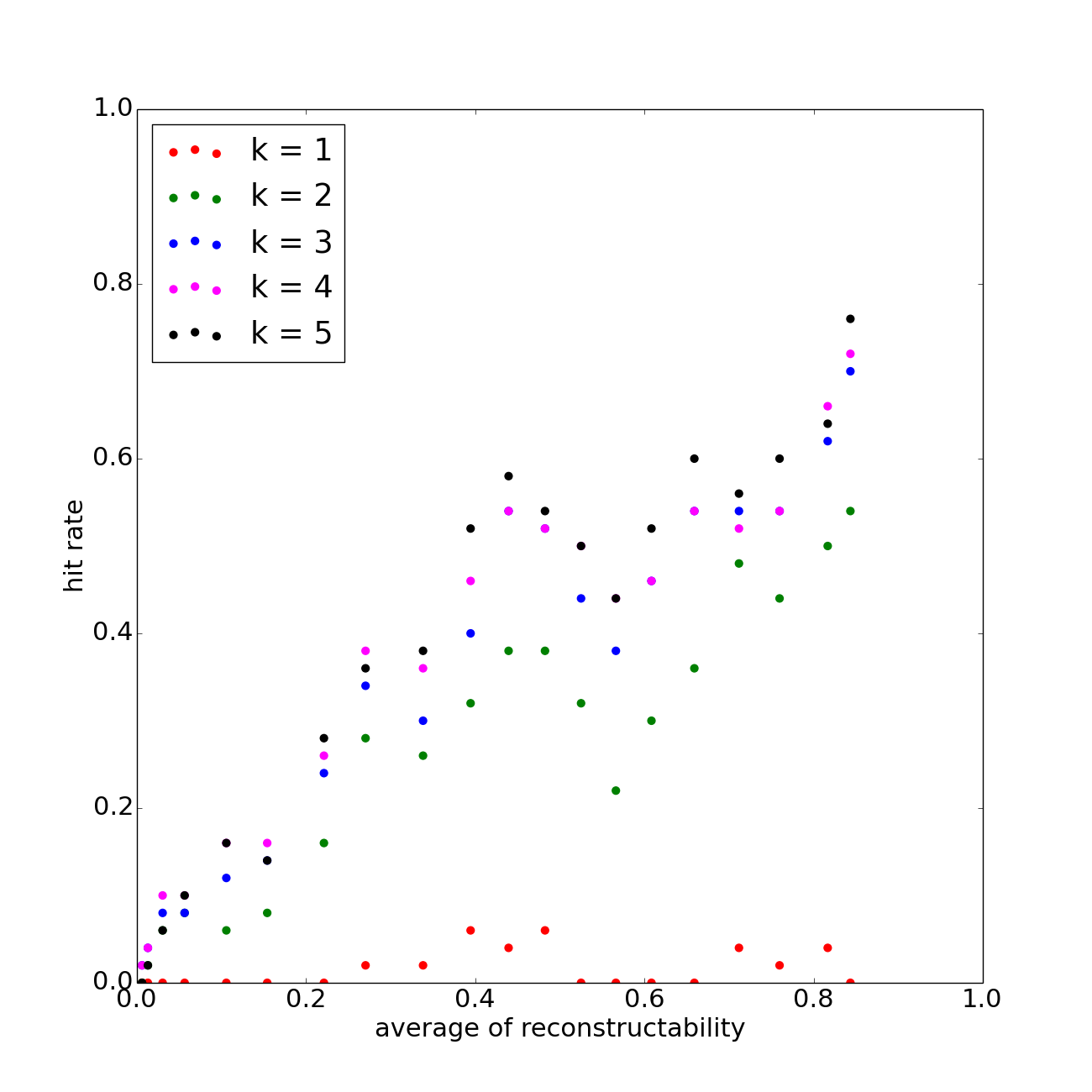}
	\end{subfigure}%
	~
	\begin{subfigure}[t]{0.33\textwidth}
		\centering
		\includegraphics[width=44mm]{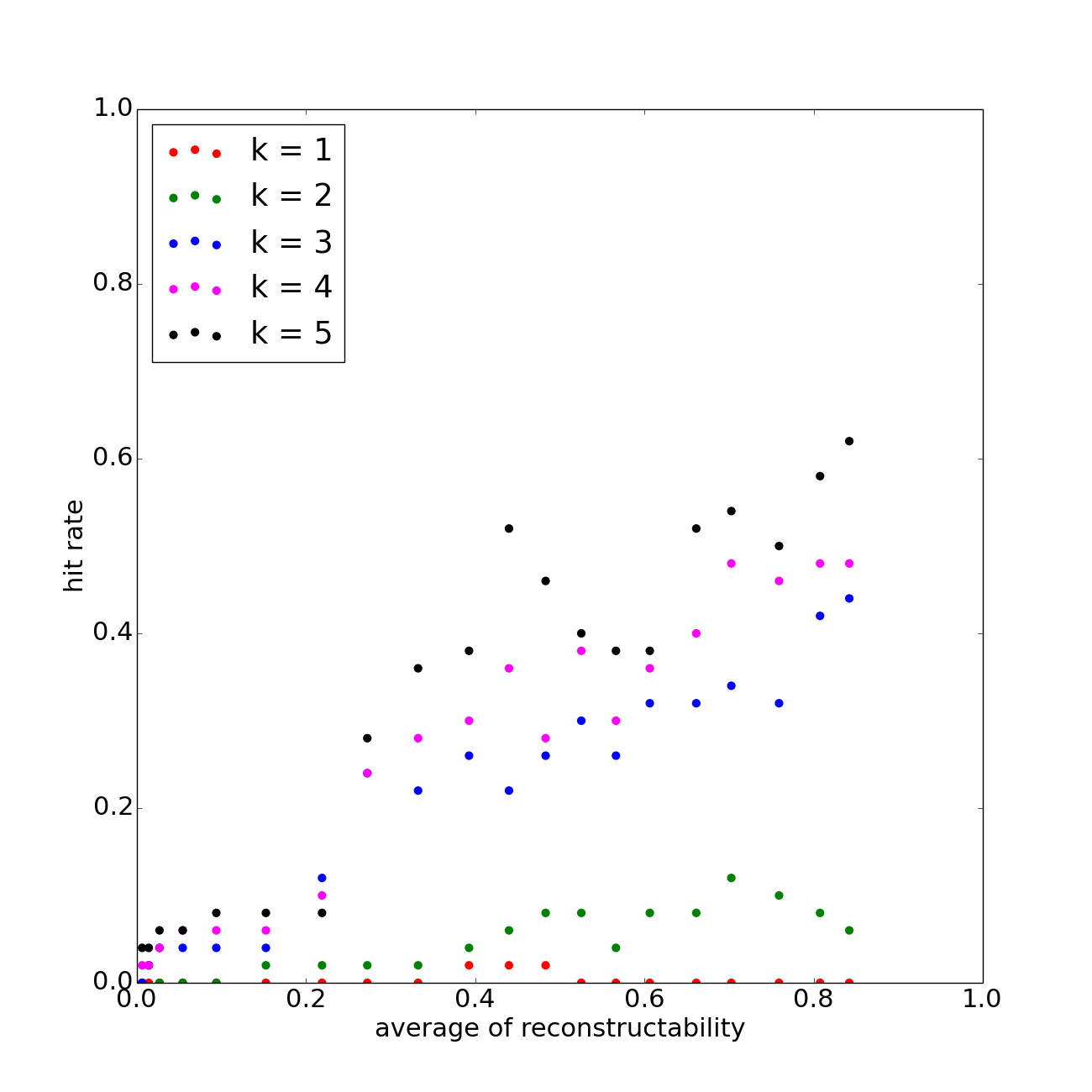}
	\end{subfigure}
	\caption{Hit-rate shown against reconstructability for $k$-means attacks with 0 (left), 60 (middle) and 120 (right) distractor terms.}
	\label{fig:trade-off} 
\end{figure*}

If the terms sent to the search engine are related to the original query, we will be able to accurately reconstruct the search results. 
However, this increases the risk of an adversary correctly guessing the query. 
Hit rate was defined as the fraction of the user queries correctly predicted by the clustering attack and is a measure of the robustness of the proposed method. 
Therefore, a natural question is \emph{what is the relationship between the reconstructability and the hit rate}.

To study this relationship, we randomly select 109 user-queries and add Gaussian noise with zero-mean and standard deviations 0 (no noise), 0.6, 1.0, 1.4 and 1.8.
In each case, we vary the number of distractor terms 0-120 and apply $k$-means clustering attacks with $k  = 1, 2, 3, 4$ and $5$.\footnote{In total, for a fixed $k$-value and the number of distractor terms, we have 545 clustering attacks.}
To conduct a conservative evaluation, we consider the terms in the vocabulary closest to the respective centroids in all clusters and not only the most coherent one as in Section~\ref{sec:attack}. 
If the original query matches any of those $k$ terms, we consider it to be a hit (e.g. to be revealing the original query).
We randomly sample data points from even intervals of reconstructability values and plot in Figure~\ref{fig:trade-off}.

We see a positive relationship between the reconstructability and the hit rate in all figures.
This indicates a trade-off between the reconstructability and the hit rate, which shows that if we try to increase the reconstructability by selecting more relevant keywords to the original user-query, then it simultaneously increases the risk of the search engine discovering the query via a clustering attack.
We see that when we increase the number of distractor terms the hit rate drops for the same value of reconstructability.
This result shows that in order to overcome the trade off between the reconstructability and the hit rate we can simply increase the number of distractor terms, thereby making the query obfuscation method more robust against clustering attacks.
Moreover, the drop due to distractor terms is more prominent for the $k=1$ attacks when we have distractor terms compared to that when we do not have distractor terms.
This is because both related and distractor terms will be contained in this single cluster from which it is difficult to guess the original user-query. 

Overall, the hit rate drops in the order $k=5$, $k=3$ and $k=2$ when we increase the number of distractor terms.
This result suggests that if one wants to increase the hit rate, then an effective strategy is to increase the number of clusters
because we consider it to be a hit if the user-query is found via any of the clusters.
However, in practice, we will need to further select one term from all the clusters.
Nevertheless, we can consider the hit rate obtained in this manner to be a more conservative estimate, whereas in reality it will be less and therefore be more robust against attacks.
We conduct a human evaluation of the distractor terms in Appendix B, which interestingly shows that the the distractor terms found by the proposed method make it hard even for humans to predict the original query.

\section{Related Work}

One of the early incidents of query logs leaking private information in the public domain is the AOL's release of query log data in 2006.\footnote{\url{https://tinyurl.com/y9qx9ufz}}
Following this incident various methods have been proposed to obfuscate user queries such as token-based hashing~\cite{Kumar:2007} and query-log bundling~\cite{Jones:CIKM:2008}.
However, in these approaches obfuscation happens only at the Web search engine's side without any intervention by the users, and the users must trust the good intentions of the search engine with respect to the user privacy.
Moreover, \cite{Kumar:2007} showed that hashing alone \emph{does not} guarantee user privacy.

Accessing Web search engines via an anonymised proxy server such as the onion routing~\cite{OnionRouting}, TOR~\cite{TOR}, Dissent~\cite{Dissent} or RAC~\cite{RAC} is a popular strategy employed by common users. 
The goal is to prevent the search engine link the queries issues by a user to his or her user profile.
Anonymised search engines such as duckduckgo, Qwant, Swisscows provide privacy-oriented alternatives to Web users where the IP addresses, search profiles, location information etc. related to the users are kept anonymised.
On the other hand, the query obfuscation method we propose in this paper can be used in conjunction with other anonymisation techniques in existing private-browsing browser plug-ins/add-ons and search portals to further increase the level of privacy.

Obfuscation-based private web search~\cite{Balsa:2012} includes dummy keywords to prevent search engines from guessing users' query intent. 
Several browser add-ons that automatically append unrelated fake terms have been developed such as TrackMetNot~\cite{TrackMeNot}, OptimiseGoogle, Google Privacy, Private Web Search tool~\cite{Saint-Jean:2007} and GooPIR~\cite{GoPIR}.
Although this approach is similar to our proposal to append user queries with distractor terms, those prior proposals
have relied on pre-compiled ontologies~\cite{Petit_2014} such as the WordNet or queries issued by other users shared via a peer network.
Such approaches have scalability issues because most named entities that appear in search queries do not appear in the WordNet and it is unlikely that users would openly share their keywords to be used by their peers.
Recently, outside IR, obfuscation has been applied successfully for anonymising users in social media platforms~\cite{masood2018incognito,papadopoulos2013k}.

The goal in Private Information Retrieval~\cite{PIR} is to retrieve data from a database without revealing the query but only some encrypted or obfuscated version of it~\cite{Ostrovsky:2007,Chor:1997}. 
For example, in hompmophic encryption-based methods the user (client) submits encrypted keywords and the search engine (server) performs a blinded lookup and returns the results again in an encrypted form, which can then be decrypted by the user. Embellishing queries with decoy terms further protects the privacy of the users. 
PIR has been applied in recommender systems~\cite{gupta2016scalable} and public data~\cite{wang2017splinter}.
 However, unlike our proposed method, PIR methods assume search engines to accommodate the client side encryption methods, which is a critical limitation because modern commercial Web search engines do not allow this.

\section{Conclusion}

We proposed a method to obfuscate queries sent to a Web search engine by decomposing the query into a set of related terms and a set of distractor terms. 
We then reconstruct the search results for the original query using the search results we obtain for the related terms, discarding the search results for the distractor terms. We theoretically studied the relationship between the obfuscity and the reconstructability obtained using the proposed method under different noise levels. 
We empirically showed that the proposed query obfuscation method is robust against a $k$-means clustering attack. 
Moreover, a human evaluation task, implemented as a query prediction game, showed that it is even difficult for humans to predict the original query from the obfuscation produced by the proposed method.
Even though the original query issued by the user can be obfuscated using the proposed method, if one or more of its related terms can be accurately discovered, it could still reveal the information intent of the user.
Therefore, we identify methods that would guarantee the obfuscation of not only the original user query, but also any of its related terms as an important future research direction.

\section*{Acknowledgements}
This project is supported by JSPS KAKENHI Grant Numbers JP18H05291 and JP20A402.

\section{Bibliographical References}
 \bibliographystyle{lrec2022-bib}
 \bibliography{black}

\begin{thebibliography}{}

\bibitem[\protect\citename{Arora \bgroup et al.\egroup }2016]{Arora:TACL:2016}
Arora, S., Li, Y., Liang, Y., Ma, T., and Risteski, A.
\newblock (2016).
\newblock A latent variable model approach to pmi-based word embeddings.
\newblock {\em Transactions of the Association for Computational Linguistics},
  4:385--399.

\bibitem[\protect\citename{Arora \bgroup et al.\egroup }2017]{Arora:ICLR:2017}
Arora, S., Liang, Y., and Ma, T.
\newblock (2017).
\newblock A simple but tough-to-beat baseline for sentence embeddings.
\newblock In {\em Proc. of ICLR}.

\bibitem[\protect\citename{Balsa \bgroup et al.\egroup }2012]{Balsa:2012}
Balsa, E., Troncoso, C., and Diaz, C.
\newblock (2012).
\newblock Ob-pws: Obfuscation-based private web search.
\newblock {\em 2012 IEEE Symposium on Security and Privacy}, May.

\bibitem[\protect\citename{Bollegala \bgroup et al.\egroup
  }2018]{Bollegala:AAAI:2018}
Bollegala, D., Yoshida, Y., and Kawarabayashi, K.-i.
\newblock (2018).
\newblock {Using $k$-way Co-occurrences for Learning Word Embeddings}.
\newblock In {\em Proc. of AAAI}, pages 5037--5044.

\bibitem[\protect\citename{Carpineto and Romano}2012]{Carpineto:2012}
Carpineto, C. and Romano, G.
\newblock (2012).
\newblock A survey of automatic query expansion in information retrieval.
\newblock {\em Journal of ACL Computing Surveys}, 44(1):1 -- 50.

\bibitem[\protect\citename{Chor \bgroup et al.\egroup }1997]{Chor:1997}
Chor, B., Gilboa, N., and Naor, M.
\newblock (1997).
\newblock Private information retrieval by keywords.
\newblock Technical report, Department of Computer Science, Technion, Israel
  Institute of Technology.

\bibitem[\protect\citename{Cordeiro \bgroup et al.\egroup
  }2016]{cordeiro-EtAl:2016:P16-1}
Cordeiro, S., Ramisch, C., Idiart, M., and Villavicencio, A.
\newblock (2016).
\newblock Predicting the compositionality of nominal compounds: Giving word
  embeddings a hard time.
\newblock In {\em Proceedings of the 54th Annual Meeting of the Association for
  Computational Linguistics (Volume 1: Long Papers)}, pages 1986--1997, Berlin,
  Germany, August. Association for Computational Linguistics.

\bibitem[\protect\citename{Corrigan-Gibbs and Ford}2010]{Dissent}
Corrigan-Gibbs, H. and Ford, B.
\newblock (2010).
\newblock Dissent: accountable anonymous group messaging.
\newblock In {\em Proc. of CCS}.

\bibitem[\protect\citename{Devlin \bgroup et al.\egroup }2019]{BERT}
Devlin, J., Chang, M.-W., Lee, K., and Toutanova, K.
\newblock (2019).
\newblock {B}{E}{R}{T}: {P}re-training of {D}eep {B}idirectional {T}ransformers
  for {L}anguage {U}nderstanding.
\newblock In {\em Proc. of NAACL-HLT}.

\bibitem[\protect\citename{Dingledine \bgroup et al.\egroup }2004]{TOR}
Dingledine, R., Mathewson, N., and Syversion, P.
\newblock (2004).
\newblock Tor: The second generation onion router.
\newblock In {\em Proc. of the Usenix Security Symposium}.

\bibitem[\protect\citename{Domingo-Ferrer \bgroup et al.\egroup }2009]{GoPIR}
Domingo-Ferrer, J., Solanas, A., and Castella-Roca, J.
\newblock (2009).
\newblock $h(k)$-private information retrieval from privacy uncooperative
  queryable databases.
\newblock In {\em Proc. of Online Information Review}, volume~33, pages
  720--744.

\bibitem[\protect\citename{Gervais \bgroup et al.\egroup }2014]{Gervais_2014}
Gervais, A., Shokri, R., Singla, A., Capkun, S., and Lenders, V.
\newblock (2014).
\newblock Quantifying web-search privacy.
\newblock {\em Proceedings of the 2014 ACM SIGSAC Conference on Computer and
  Communications Security - CCS '14}.

\bibitem[\protect\citename{Goldschlang \bgroup et al.\egroup
  }1999]{OnionRouting}
Goldschlang, D., Reed, M., and Syverson, P.
\newblock (1999).
\newblock Onion routing.
\newblock {\em Communications of the ACM}, 42(2).

\bibitem[\protect\citename{Gupta \bgroup et al.\egroup
  }2016]{gupta2016scalable}
Gupta, T., Crooks, N., Mulhern, W., Setty, S., Alvisi, L., and Walfish, M.
\newblock (2016).
\newblock Scalable and private media consumption with popcorn.
\newblock In {\em 13th $\{$USENIX$\}$ Symposium on Networked Systems Design and
  Implementation ($\{$NSDI$\}$ 16)}, pages 91--107.

\bibitem[\protect\citename{Hashimoto and Tsuruoka}2016]{Hashimoto:ACL:2016}
Hashimoto, K. and Tsuruoka, Y.
\newblock (2016).
\newblock Adaptive joint learning of compositional and non-compositional phrase
  embeddings.
\newblock In {\em Proc. of ACL}, pages 205--215.

\bibitem[\protect\citename{He \bgroup et al.\egroup }2008]{He:2008}
He, C., Wang, C., Zhong, Y.-X., and Li, R.-f.
\newblock (2008).
\newblock A survey on learning to rank.
\newblock In {\em Proc. of the 7th Intl. Conf. on Machine Learning and
  Cybernetics}, pages 1734 -- 1739.

\bibitem[\protect\citename{Howe and Nissenbaum}2009]{TrackMeNot}
Howe, D.~C. and Nissenbaum, H.
\newblock (2009).
\newblock Trackmenot: Resisting surveillance in web search.
\newblock Lessons from the Identity Train: Anonymity, Privacy and Identity in a
  Networked Society.

\bibitem[\protect\citename{Jones \bgroup et al.\egroup }2008]{Jones:CIKM:2008}
Jones, R., Kumar, R., Pang, B., and Tomkins, A.
\newblock (2008).
\newblock Vanity fair: Privacy in querylog bundles.
\newblock In {\em Proc. of CIKM}, pages 853--862.

\bibitem[\protect\citename{Kumar \bgroup et al.\egroup }2007]{Kumar:2007}
Kumar, R., Novak, J., Pang, B., and Tomkins, A.
\newblock (2007).
\newblock On anonymizing query logs via token-based hashing.
\newblock In {\em Proceedings of the 16th International Conference on World
  Wide Web}, WWW '07, pages 629--638, New York, NY, USA. ACM.

\bibitem[\protect\citename{Masood \bgroup et al.\egroup
  }2018]{masood2018incognito}
Masood, R., Vatsalan, D., Ikram, M., and Kaafar, M.~A.
\newblock (2018).
\newblock Incognito: A method for obfuscating web data.
\newblock In {\em Proceedings of the 2018 World Wide Web Conference}, pages
  267--276. International World Wide Web Conferences Steering Committee.

\bibitem[\protect\citename{Mikolov \bgroup et al.\egroup }2013]{Milkov:2013}
Mikolov, T., Chen, K., and Dean, J.
\newblock (2013).
\newblock Efficient estimation of word representation in vector space.
\newblock In {\em Proc. of International Conference on Learning
  Representations}.

\bibitem[\protect\citename{Miller}1995]{WordNet}
Miller, G.~A.
\newblock (1995).
\newblock Wordnet: A lexical database for english.
\newblock {\em Communications of the ACM}, 38(11):39 -- 41, November.

\bibitem[\protect\citename{Mokhtar \bgroup et al.\egroup }2013]{RAC}
Mokhtar, S.~B., Berthou, G., Diarra, A., Qu{\'e}ma, V., and Shoker, A.
\newblock (2013).
\newblock Rac: A freerider-resilient scalable, anonymous communication
  protocol.
\newblock In {\em Proc. of ICDCS}.

\bibitem[\protect\citename{Ostrovsky and Skeith}2007]{Ostrovsky:2007}
Ostrovsky, R. and Skeith, W.~I.
\newblock (2007).
\newblock A survey of single-database pir: techniques and applications.
\newblock In {\em Proc. of Public Key Cryptography (PKC)}, volume 4450, pages
  393--411.

\bibitem[\protect\citename{Papadopoulos \bgroup et al.\egroup
  }2013]{papadopoulos2013k}
Papadopoulos, P., Papadogiannakis, A., Polychronakis, M., Zarras, A., Holz, T.,
  and Markatos, E.~P.
\newblock (2013).
\newblock K-subscription: Privacy-preserving microblogging browsing through
  obfuscation.
\newblock In {\em Proceedings of the 29th Annual Computer Security Applications
  Conference}, pages 49--58. ACM.

\bibitem[\protect\citename{Pasca}2007]{Pasca:WWW:2007}
Pasca, M.
\newblock (2007).
\newblock Organizing and searching the world wide web of facts-step two:
  Harnessing the wisdom of the crowds.
\newblock In {\em WWW 2007}, pages 101--110.

\bibitem[\protect\citename{Pasca}2014]{pasca:2014:EMNLP2014}
Pasca, M.
\newblock (2014).
\newblock Queries as a source of lexicalized commonsense knowledge.
\newblock In {\em Proc. of EMNLP}, pages 1081--1091.

\bibitem[\protect\citename{Pennington \bgroup et al.\egroup
  }2014]{Pennington:EMNLP:2014}
Pennington, J., Socher, R., and Manning, C.~D.
\newblock (2014).
\newblock Glove: global vectors for word representation.
\newblock In {\em Proc. of EMNLP}, pages 1532--1543.

\bibitem[\protect\citename{Peters \bgroup et al.\egroup }2018]{Elmo}
Peters, M.~E., Neumann, M., Iyyer, M., Gardner, M., Clark, C., Lee, K., and
  Zettlemoyer, L.
\newblock (2018).
\newblock {D}eep contextualized word representations.
\newblock In {\em Proc. of NAACL-HLT}.

\bibitem[\protect\citename{Petit \bgroup et al.\egroup }2014]{Petit_2014}
Petit, A., Ben~Mokhtar, S., Brunie, L., and Kosch, H.
\newblock (2014).
\newblock Towards efficient and accurate privacy preserving web search.
\newblock {\em Proceedings of the 9th Workshop on Middleware for Next
  Generation Internet Computing - MW4NG '14}.

\bibitem[\protect\citename{Poliak \bgroup et al.\egroup
  }2017]{Poliak:EACL:2017}
Poliak, A., Rastogi, P., Martin, M.~P., and Durme, B.~V.
\newblock (2017).
\newblock Efficient, compositional, order-sensitive $n$-gram embeddings.
\newblock In {\em Proc. of EACL}.

\bibitem[\protect\citename{Richardson}2008]{Richardson:TWEB:2008}
Richardson, M.
\newblock (2008).
\newblock Learning about the world through long term query logs.
\newblock {\em ACM Transactions on the Web}, 2(4).

\bibitem[\protect\citename{Sadikov \bgroup et al.\egroup
  }2010]{Sadikov:WWW:2010}
Sadikov, E., Madhavan, J., Wang, L., and Halevy, A.
\newblock (2010).
\newblock Clustering query refinements by user intent.
\newblock In {\em WWW 2010}, pages 841--850.

\bibitem[\protect\citename{Saint-Jean \bgroup et al.\egroup
  }2007]{Saint-Jean:2007}
Saint-Jean, F., Johnson, A., Boneh, D., and Feigenbaum, J.
\newblock (2007).
\newblock Private web search.
\newblock In {\em Proc. of ACM Workshop on Privacy in Electronic Society},
  pages 84--90.

\bibitem[\protect\citename{Santos \bgroup et al.\egroup }2010]{Santos:WWW:2010}
Santos, R. L.~T., Macdonald, C., and Ounis, I.
\newblock (2010).
\newblock Exploiting query reformulations for web search result
  diversification.
\newblock In {\em WWW 2010}, pages 881--890.

\bibitem[\protect\citename{Turney and Pantel}2010]{Turney:JAIR:2010}
Turney, P.~D. and Pantel, P.
\newblock (2010).
\newblock From frequency to meaning: Vector space models of semantics.
\newblock {\em Journal of Aritificial Intelligence Research}, 37:141 -- 188.

\bibitem[\protect\citename{Wang \bgroup et al.\egroup }2017]{wang2017splinter}
Wang, F., Yun, C., Goldwasser, S., Vaikuntanathan, V., and Zaharia, M.
\newblock (2017).
\newblock Splinter: Practical private queries on public data.
\newblock In {\em 14th $\{$USENIX$\}$ Symposium on Networked Systems Design and
  Implementation ($\{$NSDI$\}$ 17)}, pages 299--313.

\bibitem[\protect\citename{Yekhanin}2010]{PIR}
Yekhanin, S.
\newblock (2010).
\newblock Private information retrieval.
\newblock {\em Commun. ACM}, 53(4):68--73, April.

\bibitem[\protect\citename{Yu and Dredze}2015]{TACL586}
Yu, M. and Dredze, M.
\newblock (2015).
\newblock Learning composition models for phrase embeddings.
\newblock {\em Transactions of the Association for Computational Linguistics},
  3:227--242.

\end{thebibliography}

 \section*{Appendix}
\appendix

\section{Proof of Theorem 1}
\label{sec:relationship}

In this section, we derive the relationship between obfuscity and reconstructability.
Because obfuscity can be increased arbitrarily by increasing the distractor terms, in this analysis, we ignore distractor terms.
This can be seen as a lower-bound for the obfuscity that  can be obtained, without using any distractor terms.
We first discuss the case where we have only one related term (i.e. $n = l = 1$) and then consider $l > 1$ reconstructability case.

\subsection{$n = l = 1$ case}
\label{sec:one-case}

Let us consider the case where $n = 1$. Here, for a given query $A$, we have only a single related term $X = X_{1}$.
In this case, $l = 1$, and we consider all documents retrieved using $X$ as relevant for $A$.
We first note that reconstructability, $\rho$, can be written as,
\begin{align}
\label{eq:l1}
\rho = \frac{|\cD(A) \cap \cD'(A)|}{|\cD(A)|}
\end{align}
from the definition of reconstructability.

Because we have a single noisy related term $X$, we have $\cD'(A) = \cD(X)$. By substituting this in \eqref{eq:l1}, we get
\begin{align}
 \label{eq:l2}
 \rho = \frac{|\cD(A) \cap \cD(X)|}{|\cD(A)|} .
\end{align}
If we consider the co-occurrence context of two terms to be the document in which they co-occur, and divide the numerator and denominator in \eqref{eq:l2} by the total number of documents indexed by the search engine, then \eqref{eq:l2} can be written as a conditional probability as in \eqref{eq:l3}.
\begin{align}
 \label{eq:l3}
 \rho = \frac{p(A,X)}{p(A)} = p(X|A)
\end{align}

Theorem 2.2 in~\cite{Arora:TACL:2016} provides a useful connection between the probability of a word (or the joint probability of
two words) and their word representations, which we summarise below.
\begin{align}
\label{eq:joint-norm}
\log p(A,X) &= \frac{\norm{v(A) + v(X)}_{2}^{2}}{2d} - 2\log Z \pm \epsilon \\
\label{eq:marginal-norm}
\log p(A) &= \frac{\norm{v(A)}_{2}^{2}}{2d} - \log Z \pm \epsilon
\end{align}
Here, $Z$ is the partition function and $\epsilon$ is the approximation error.
\eqref{eq:marginal-norm}  shows the relationship between the norm of  the embedding of a word and the frequency of that word in a corpus, whereas \eqref{eq:joint-norm} shows the relationship between the norm of the addition of the embeddings of two words and the co-occurrence frequency of those two words in a corpus. Both these relations are proved by \newcite{Arora:TACL:2016} and we would like to direct the interested readers to the original paper for the detailed proofs. 

Next, by taking the logarithm of both sides in \eqref{eq:l3} we obtain,
\begin{align}
\log \rho &= \log p(A,X) - \log p(A) \nonumber \\
\label{eq:rho-one}
&= \frac{\norm{v(X)}_{2}^{2} + 2v(X)\T v(A)}{2d} - \log Z 
\end{align}
Obfuscity for a single query term $X$ can be computed using the cosine similarity as follows:
\begin{align}
\label{eq:alpha-one}
\alpha = 1 - \frac{v(A) \T v(X)}{\norm{v(A)}_{2} \norm{v(X)}_{2}}
\end{align}
By substituting \eqref{eq:alpha-one} in \eqref{eq:rho-one} we get,

{\small
\begin{align}
\label{eq:rel}
\log \rho = \frac{\norm{v(X)}_{2}^{2}}{2d} + \frac{(1-\alpha)\norm{v(A)}_{2} \norm{v(X)}_{2}}{d} -\log Z.
\end{align}
}
Because $A$ is a given query, $v(A)$ is a constant. Moreover, if we assume that different related terms $X_{i}$ have similar
norms, (from \eqref{eq:marginal-norm} it follows that such related terms must have similar frequencies of occurrence in the corpus),
then from \eqref{eq:rel} we see that there exists a linear inverse relationship between $\log \rho$ and $\alpha$.
Because logarithm function is monotonically increasing, \eqref{eq:rel} implies an inverse relationship between $\rho$ and $\alpha$.

\subsection{$n = l > 1$ case}
\label{sec:k-case}

Let us now extend the relationship given by \eqref{eq:rel} to the case where we consider
a document to be relevant if it can be retrieved from all of the $n$ related terms.
 In other words, we have $l = n$ reconstructability in this case.
Because each search result is retrieved by all $l$ terms, we have
\begin{align}
\label{eq:results}
\cD'(A) = \cap_{i=1}^{l} \cD(X_{i}) .
\end{align}

\begin{figure*}[t]
	\centering
	\includegraphics[width=150mm]{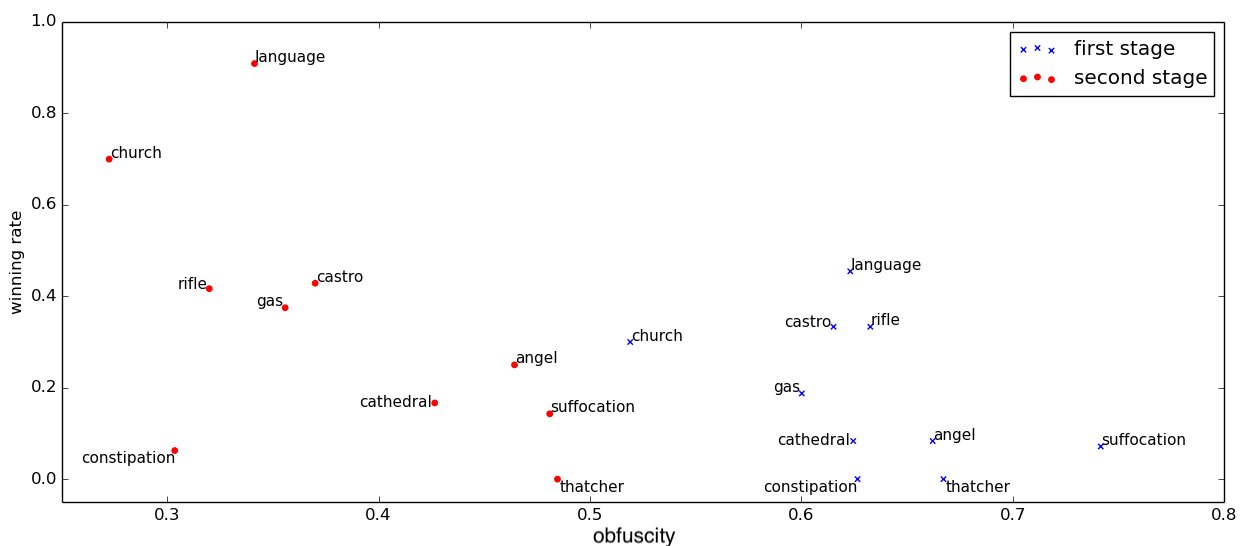}
	\caption{Winning rate vs. obfuscity for the first and second stages of the query prediction game}
	\label{fig:game}
\end{figure*}

Reconstructability can be computed in this case as follows:
\begin{align}
\label{eq:approx}
\rho &= \frac{p(A,X_{1}, X_{2}, \ldots, X_{k})}{p(A)} \nonumber \\
&= p(X_{1}, X_{2}, \ldots, X_{l} | A) \nonumber \\
& \approx   \prod_{i=1}^{l} p(X_{i}|A) 
\end{align}
In \eqref{eq:approx} we have assumed that the related terms are mutually independent given the query $A$.

Let us take the logarithm on both sides of \eqref{eq:approx}, and use \eqref{eq:joint-norm} and \eqref{eq:marginal-norm} in the same manner as we did in Section~\ref{sec:one-case} to derive the relationship given by \eqref{eq:k1}.
\begin{align}
\label{eq:k1}
\log \rho =& \frac{1}{2d}\sum_{i=1}^{l} \norm{v(X_{i})}_{2}^{2} + \nonumber \\
 &\frac{1}{d}\sum_{i=1}^{l} v(A)\T v(X_{i}) - \log Z
\end{align}

In the $n=l$ case, obfuscity can be computed as follows:
\begin{align}
\label{eq:k-anon}
\alpha = 1 - \frac{1}{l} \sum_{i=1}^{l} \frac{v(A)\T v(X_{i})}{\norm{v(A)}_{2} \norm{v(X_{i})}_{2}} 
\end{align}

Let us further assume that all related terms $X_{1}, X_{2}, \ldots, X_{l}$ occur approximately the same number of times in the corpus.
From \eqref{eq:marginal-norm} it then follows that $\norm{v(X_{i})}_{2} = c$ for $i=1,2, \ldots, l$ for some $c \in \R$.
By plugging \eqref{eq:k-anon} in \eqref{eq:k1}, and using the approximation $\norm{v(X_{i})}_{2} = c$ we arrive at the relationship between $\rho$, $\alpha$, and $l$ given by \eqref{eq:rel-k}.
\begin{align}
\label{eq:rel-k}
\log \rho = \frac{cl}{2d} \left( c + 2(1-\alpha)\norm{v(A)}_{2} \right) - \log Z & \\ \nonumber
\qed &
\end{align}

\subsection{General Case}
In the general case of $l$-reconstructability, we will have a subset of $l \leq n$ related terms retrieving each document.
Exact analysis of this case is hard, and the reconstructability given by \eqref{eq:rel-k} must be considered as a lower-bound for this general case because we will still be able to reconstruct the search results using $n \choose l$ subsets of $l$ related terms selected from a set of $n$ related terms.
Obfuscity can be arbitrarily increased without affecting the reconstructability by simply increasing the number of distractor terms.
However, doing so would increase the burden on the search engine and is not recommended.
In our experiments, we find that 20-40 distractor terms to be adequate to provide a good balance between obfuscity and efficiency.

The theoretical analysis presented in Section~\ref{sec:relationship} depends on the relationships given by \eqref{eq:joint-norm} and \eqref{eq:marginal-norm} for joint and marginal probabilities of unigram co-occurrences, originally proved by \newcite{Arora:TACL:2016}.
However, these relationships were later extended to cover co-occurrences of higher-order $n$-grams by~\newcite{Bollegala:AAAI:2018}, who showed that the squared sum of embeddings of constituent unigrams in an $n$-gram phrase is proportional to the logarithm of the joint probability of those unigrams. 
On the other hand, \newcite{Arora:ICLR:2017} showed that the inverse frequency-weighted average of unigram embeddings to be a competitive alternative to word-order sensitive supervised recurrent models for the purpose of creating phrase embeddings.
Therefore, the relationship given in \eqref{eq:rel-k} still holds both theoretically and empirically in the case of multi-word queries, enabling us to extend the proposed method to multi-word phrasal queries.

\section{Human Evaluation}
\label{sec:human}

To empirically evaluate the difficulty, not only for a search engine as done in previous sections, but even also for human attackers to predict the original query given the related and distractor terms, we devise a \emph{query prediction game}, where a group of human attackers are required to predict the original query from the related and distractor terms suggested by the proposed method. 
A group of 63 graduate students (all native English speakers) participated in this experiment.
The query prediction game is conducted in two stages. In the first stage, we randomly shuffle the related and distractor term sets extracted by the proposed method for a hidden query. 
Human attackers are unaware as to which of the terms are related to the original user-query and which are distractors.
A human attacker has a single guess to predict the user-query and wins only if the original query is correctly predicted.
If the human attacker fails at this first step, then we remove all distractor terms and display only the related terms to the human attacker.
This is expected to significantly simplify the prediction task because now the candidate set is smaller, does not contain distractor terms and the human attacker has already had a shot at the prediction.
The human attacker then has a second chance to predict the original query from the related set of terms.
If the human attacker correctly predicts the original query in the second stage, we consider it to be a winning case. 
Otherwise, the current round of the game is terminated and the next set of terms are shown to the human attacker.
Winning rate is defined as the number of games won by the human attackers, where the original user query was correctly predicted.

Figure~\ref{fig:game} shows the winning rates  for the first and second stages of the query prediction game against the obfuscity of the queries.
All queries selected for the prediction game have reconstructability scores greater than 0.3, which indicates that the search results for the original query can be accurately reconstructed from the related terms shown to the human attackers.
From Figure~\ref{fig:game}, we see that the winning rate for the first stage is lower than that for the second stage, indicating that it is easier for humans to guess the original query when the distractor terms are removed. 
Moreover, we see a gradual negative correlation between hit rate and obfuscity. 
This shows that more obfuscatory the terms are, it becomes difficult even for the human attackers to predict the original query, which is a desirable property for a query obfuscation method.

 \end{document}